\documentclass[10pt,a4paper,twosides]{article}

\usepackage[table,xcdraw]{xcolor}
\usepackage{color}
\usepackage{empheq}

\usepackage{geometry}
\usepackage{float}
\usepackage{subfiles}
\usepackage{xr}
\usepackage{amsmath,amssymb,amsfonts}
\usepackage{mathtools}
\usepackage{mathrsfs}
\usepackage{makeidx}

\usepackage[textsize=footnotesize]{todonotes} 
\usepackage{exscale,relsize} 
\usepackage{mleftright}
\usepackage{theorem}
\usepackage{graphics}                 
\usepackage{accents}
\usepackage{kbordermatrix}
\usepackage[margin=10pt,font=small,labelfont=bf]{caption}
\usepackage{tikz}
\tikzstyle{every picture}=[scale=0.75,transform shape]
\usepackage{verbatim}
\usetikzlibrary{arrows,automata,arrows.meta}
\usepackage{cases}
\usepackage{makeidx}
\usepackage{longtable}
\usepackage{tabu}
\usepackage{algorithmic}
\usepackage{algorithm}
\usepackage{caption}
\usepackage{subfigure}
\usepackage{url}
\usepackage{bm,xstring}

\allowdisplaybreaks

\def\expm{\mathop{\mathrm{expm}}}

\newcommand{\mat}[1]{\mathbf{#1}}

\newcommand{\argmax}{\mathop{\mathrm{arg\,max}}}
\newcommand{\minimize}{\mathop{\mathrm{minimize}}}

\newcommand{\subjectto}{\mathop{\mathrm{subject\,to}}}

\newcommand{\myDelta}{{\textstyle \mathsmaller{\varDelta}}} 

\newcommand{\dsum}{\displaystyle\sum}

\newcommand{\cditto}{\raisebox{.5ex}{\hbox to 2em{\hrulefill}}}

\newcommand{\bigzero}{\mbox{\normalfont\large\bfseries 0}}
\newcommand{\mypound}{\scalebox{0.8}{\raisebox{0.4ex}{\#}}}

\def\expm{\mathop{\mathrm{expm}}}

\mleftright 

\title{Randomized Shortest Paths with\\
Net Flows and Capacity Constraints\\
\quad \\
  \normalsize{\textit{To appear in
Information Sciences}}}

\author{Sylvain Courtain, Pierre Leleux, Ilkka Kivimaki, \\
Guillaume Guex \& Marco Saerens}

\begin{document}

\setlength{\parskip}{1pt plus 1pt minus 1pt} 

\sloppy 

\date{}
\maketitle

\begin{abstract} 
This work extends the randomized shortest paths (RSP) model by investigating the net flow RSP and adding capacity constraints on edge flows. The standard RSP is a model of movement, or spread, through a network interpolating between a random-walk and a shortest-path behavior \cite{Kivimaki-2012,Saerens-2008,Yen-08K}. The framework assumes a unit flow injected into a source node and collected from a target node with flows minimizing the expected transportation cost, together with a relative entropy regularization term. In this context, the present work first develops the net flow RSP model considering that edge flows in opposite directions neutralize each other (as in electric networks), and proposes an algorithm for computing the expected routing costs between all pairs of nodes. This quantity is called the net flow RSP dissimilarity measure between nodes. Experimental comparisons on node clustering tasks indicate that the net flow RSP dissimilarity is competitive with other state-of-the-art dissimilarities. In the second part of the paper, it is shown how to introduce capacity constraints on edge flows, and a procedure is developed to solve this constrained problem by exploiting Lagrangian duality. These two extensions should improve significantly the scope of applications of the RSP framework.
\end{abstract}

\section{Introduction}
\label{Sec_introduction01}

Link analysis and network science are dedicated to the analysis of network data and are currently studied in a large number of different fields (see, e.g., \cite{Newman-2018}).
One recurring problem in this context is the definition of meaningful measures for capturing interesting properties of the network and its nodes, like distance measures between nodes or centrality measures. These quantities usually take the structure of the whole network into account.

However, many such measures are based on strong assumptions about the movement, or communication, occurring in the network whose two extreme cases are the optimal behavior and the random behavior. Indeed, the two most popular distance measures in this context are the least-cost distance (also called shortest-path distance) and the resistance distance \cite{Klein-1993} (equal to the effective resistance and proportional to the commute-time distance when considering a random walk on the graph; see \cite{FoussKDE-2005} and references therein).
The same holds with the betweenness centrality: popular measures are the shortest-path betweenness introduced by Freeman \cite{Freeman-1977} and the random-walk betweenness (also called the current-flow betweenness) independently introduced by Newman \cite{Newman-05} and by Brandes and Fleischer \cite{Brandes-2005b}.
Still another example is provided by the measures of compactness of a network, such as the Wiener index (based on shortest paths) and the Kirchhoff index (based on random walks or electric networks).

In reality, however, behavior is seldom based on complete randomness or optimality. Therefore, a large effort has been invested in defining models interpolating between a shortest-path and a random-walk behavior \cite{Fouss-2016}, especially in the context of distance measures between nodes where both proximity and  high connectivity are taken into account (the concept of relative accessibility \cite{Chebotarev-1997}). These models are based on extensions of electric networks \cite{vonLuxburg-2011,Herbster-2009,Nguyen-2016}, on combinatorial analysis arguments \cite{Chebotarev-2011,Chebotarev-2012,Chebotarev-1997}, on mixed L1-L2 regularization \cite{Li-2013}, on entropy-regularized network flow models \cite{Bavaud-2012,Guex-2016,Guex-2015}, or on entropy-regularized least-cost paths (\cite{Kivimaki-2012,Saerens-2008,Yen-08K}, directly inspired by a transportation model, denoted as the Markovian traffic assignment in this field \cite{Akamatsu-1996}, and also related to \cite{Todorov-2007}). The latter (i.e., the  entropy-regularized least-cost paths) constitutes the \textbf{randomized shortest paths} (RSP) framework which is the main subject of this paper.
For a more thorough discussion of families of distances between nodes, see, for example, \cite{Fouss-2016,Kivimaki-2012}.

This effort is pursued here by proposing two extensions to this RSP model, considering:
\begin{itemize}
\item A new dissimilarity measure extending the RSP dissimilarity \cite{Kivimaki-2012,Saerens-2008,Yen-08K}, namely the \textbf{net-flow RSP dissimilarity}. Like the standard RSP dissimilarity, this new dissimilarity is the expected cost needed for reaching the target node from the source node, but now considering that edge flows in two opposite directions cancel each other out, as for the electric current \cite{Snell-1984}. Therefore, this model of movement based on net flows more resembles electric flows when the temperature of the system is high. An algorithm is proposed for computing the net-flow RSP dissimilarity matrix between all pairs of nodes.
\item The introduction of \textbf{capacity constraints} in the model. Capacity constraints on edge flows are very common in practice \cite{Ahuja-1993}, and the applicability of the RSP model would certainly be increased if such constraints could be integrated.
Therefore, the main contributions related to capacity constraints are (1) to show how the model can accommodate such constraints, for both raw edge flows and net flows, and (2) to provide an algorithm for solving the constrained RSP model in the case of a single pair of source/target nodes.
\end{itemize}

Capacity constraints appear frequently in network flow problems, and there is a vast literature on the subject, especially in the operations research field (see, e.g., \cite{Ahuja-1993,Korte-2018}). They arise, for instance, in the standard \emph{minimum cost flow} problem that aims to find the source-target flow with minimal cost and subject to some capacity constraints. As discussed in \cite{Ahuja-1993}, this task often appears in real-world problems. In short, capacity constraints are mainly present in order to, for example, avoid congestion, spreading the traffic, or simply because the flow is restricted. In this context, various algorithms for solving the standard minimum cost flow problem have been proposed: minimum mean cycle-cancelling, successive shortest paths, network simplex, and primal-dual, to name a few (see the above citations and references therein). The difference with our work is that, in the RSP, (1) the model is expressed in terms of full paths and (2) a Kullback-Leibler regularization term is introduced, inducing randomized routing policies that can be computed by standard matrix operations.

In the transportation networks field that inspired the RSP \cite{Akamatsu-1996,Dial71}, capacity constraints were also considered in various transportation models, such as the standard traffic assignment problem \cite{Hearn-1990}, the deterministic static equilibrium model \cite{Ferrari-1995}, the stochastic user equilibrium model \cite{Bell-1995}, or the random utility theory \cite{Ryu-2014} (see also the references in these papers).

Among them, the closest to our work is the stochastic user equilibrium model.
However, to the best of our knowledge, we are not aware of such models integrating capacity constraints based on an RSP-type full paths formalism.
Therefore, because of their practical usefulness, we found that it would be a valuable contribution to introduce such constraints within the framework, which is developed in Section \ref{Sec_edge_flow_capacity_constraints01}. Indeed, we will exploit the fact that, as the RSP model can be derived using maximum entropy arguments \cite{Cover-2006,Jaynes-1957}, linear inequality constraints can be handled by Lagrangian duality (see, e.g., \cite{Jebara-2004}).

The content of this paper is structured as follows. Section \ref{Sec_randomized_shortest_paths01} summarizes the standard RSP framework. Section \ref{Sec_net_flow_RSP01} introduces the net flow RSP dissimilarity. Then, Section \ref{Sec_edge_flow_capacity_constraints01} develops new algorithms for constraining the flow capacity on edges, while Section \ref{Sec_net_flow_capacity_constraints01} deals with net flow capacity constraints. Illustrative examples and experiments on node clustering tasks are described in Section \ref{Sec_experiments01}. Finally, Section \ref{Sec_conclusion01} presents the conclusion.

\section{The standard randomized shortest paths framework}
\label{Sec_randomized_shortest_paths01}

As already discussed, the main contributions of the paper are based on the RSP framework, interpolating between a least-cost and a random-walk behavior, and allowing dissimilarity measures to be defined between nodes \cite{Kivimaki-2012,Saerens-2008,Yen-08K}. The formalism, inspired by models developed in transportation science \cite{Akamatsu-1996,Dial71}, is based on full paths instead of standard ``local" edge flows \cite{Ahuja-1993,Bavaud-2012,Guex-2015} and is briefly described in this section for completeness.
We start by providing a short account of the RSP before introducing, in the following sections, the net flow RSP dissimilarity as well as the algorithm for solving the flow-capacity constrained RSP problem on a directed graph.

\subsection{Background and notation}

Let us begin by introducing some necessary notation \cite{Fouss-2016,Kivimaki-2012}. First, notice that column vectors are written in bold lowercase and matrices are in bold uppercase.
Moreover, in this work, we consider a weighted directed,\footnote{When the graph is undirected, we consider that it is made of two reciprocal, directed, edges with the same affinities and costs.} strongly connected graph or network, $G = (\mathcal{V}, \mathcal{E})$, with a set $\mathcal{V}$ of $n$ nodes and a set $\mathcal{E}$ of $m$ directed edges. An edge connecting node $i$ to node $j$ is denoted by $(i,j)$ or $i \rightarrow j$.
Furthermore, we are given an \textbf{adjacency matrix} $\mathbf{A} = (a_{ij}) \ge 0$ quantifying the directed, local, positive affinity between pairs of adjacent nodes $i$, $j$. As usual, a zero value, $a_{ij} = 0$, indicates that there is no edge between nodes $i$ and $j$. In addition, we assume that there are no self-loops in the network, that is, $a_{ii}=0$ for all $i$.
From this adjacency matrix, the standard \textbf{reference random walk} on the graph is defined in
the usual way: the \textbf{transition probabilities} associated with each node are set proportionally to the affinities and then normalized in order to sum to one,
\begin{equation}
p_{ij}^{\mathrm{ref}} = \frac{a_{ij}}{{\sum_{j'=1}^{n}} a_{ij'}}
= \frac{a_{ij}}{d_{i}}
\label{Eq_Transition_probabilities01}
\end{equation}
where $d_{i}$ is the (out)degree of node $i$.
The matrix $\mathbf{P}_{\mathrm{ref}} = (p_{ij}^{\mathrm{ref}})$ is row-stochastic and is called the transition matrix of the natural, reference, random walk on the graph. 

In addition, a transition \textbf{cost}, $c_{ij} \ge 0$, is
associated with each edge $(i,j)$ of $G$. If there is no
edge linking $i$ to $j$, the cost is assumed to take an infinite value,
$c_{ij} = \infty$. We assume for consistency that $c_{ij} = \infty$ if $a_{ij} = 0$, and
the cost matrix is defined accordingly, $\mathbf{C} = (c_{ij})$.
Costs are usually set independently of the adjacency matrix; they quantify the immediate penalty associated with a transition, depending on the application at hand. This is, however, not always the case. For example, in electric networks, the costs are resistances and the affinities are conductances: in this context, they are linked by $a_{ij}=1/c_{ij}$.

A \textbf{path} or \textbf{walk} $\wp$ is a finite sequence of hops to adjacent nodes on $G$ (including cycles), initiated from a source node $s$ and stopping at some ending target node $t$ with $s \ne t$. A \textbf{hitting path} is a path where the last node $t$ does not appear as an intermediate node. In other words, a hitting path to node $t$ stops when it reaches $t$ for the first time. In practice, we consider hitting paths to the fixed target node $t$ by setting this target node as absorbing (or ``killing"). Computationally, this is achieved by putting the corresponding row $t$ of the transition matrix to zero. 
The node at position $\tau$ along path $\wp$ is denoted by $\wp(\tau)$. The \textbf{total cost} of a path, $\tilde{c}(\wp)$, is simply the sum of the edge costs $c_{ij}$ along $\wp$, while its \textbf{length} $\ell(\wp)$ is the number of steps, or hops, needed for following that entire path.

\subsection{The standard randomized shortest paths formalism}

The main idea behind the RSP model is closely related to maximum entropy problems \cite{Cover-2006,Jaynes-1957}. Let us consider the set of all hitting paths, or walks, $\wp \in \mathcal{P}_{st}$ from a node $s \in \mathcal{V}$ (source) to an absorbing, killing node $t \in \mathcal{V}$ (target) on $G$. Then, we assign a probability distribution $\mathrm{P}(\cdot)$ on the discrete set of paths $\mathcal{P}_{st}$ \cite{Bavaud-2012,Kivimaki-2012} by minimizing the \textbf{free energy} of statistical physics \cite{Jaynes-1957}, \footnote{More precisely, it corresponds to a generalized free energy based on the relative entropy instead of the Shannon entropy.}
\begin{equation}
\vline\,\begin{array}{llll}
\minimize\limits_{\{ \mathrm{P}(\wp) \}_{\wp \in \mathcal{P}_{st}}} & \phi(\mathrm{P}) = \dsum_{\wp \in \mathcal{P}_{st}} \mathrm{P}(\wp) \tilde{c}(\wp) + T \dsum_{\wp \in \mathcal{P}_{st}} \mathrm{P}(\wp) \log \left( \frac{\mathrm{P}(\wp)}{\tilde{\pi}(\wp)} \right) \\[0.5cm]
\subjectto & \sum_{\wp\in\mathcal{P}_{st}}\textnormal{P}(\wp) = 1
\end{array}
\label{Eq_optimization_problem_BoP01}
\end{equation}
where $\tilde{c}(\wp) = \sum_{\tau = 1}^{\ell(\wp)} c_{\wp(\tau-1) \wp(\tau)}$
is the total cumulated cost along path $\wp$ when visiting the nodes $\left( \wp(\tau) \right)_{\tau=0}^{\ell(\wp)}$ in the sequential order. Furthermore, $\tilde{\pi}(\wp) = \prod_{\tau = 1}^{\ell(\wp)} p_{\wp(\tau-1) \wp(\tau)}^{\mathrm{ref}}$
is the product of the reference transition probabilities along path $\wp$, i.e., the random walk probability of path $\wp$.

The objective function (Eq.\ \ref{Eq_optimization_problem_BoP01}) is a mixture of two dissimilarity terms, with the temperature $T>0$ balancing the trade-off between them.
The first term is the expected cost for reaching the target node from the source node (favoring shorter paths -- \emph{exploitation}). The second term corresponds to the relative entropy \cite{Cover-2006}, or Kullback-Leibler divergence, between the path probability distribution and the reference (random-walk) paths probability distribution (introducing randomness and diversity -- \emph{exploration}). For a low temperature $T$, low-cost paths are favored, whereas when $T$ is large, paths are chosen according to their likelihood in the reference random walk on $G$.

The problem (\ref{Eq_optimization_problem_BoP01}) corresponds to a standard minimum cost flow problem, as discussed in the introduction,\footnote{Here, without capacity constraints, which will be introduced in Section \ref{Sec_edge_flow_capacity_constraints01}.} with a Kullback-Leibler regularization term expressed in terms of full paths in the RSP formalism. There are, however, some subtle differences, such as the fact that in the standard minimum cost flow problem the flows are unidirectional, whereas the RSP defines a Markov chain for which flows are generally bi-directional.

This argument, akin to maximum entropy \cite{Cover-2006,Jaynes-1957}, leads to a \textbf{Gibbs-Boltzmann distribution} on the set of paths (see, e.g., \cite{Kivimaki-2012, Saerens-2008}),
\begin{equation}
\mathrm{P}^{*}(\wp) 
= \frac{\tilde{\pi}(\wp) \exp[-\theta \tilde{c}(\wp)]}{\dsum_{\wp'\in\mathcal{P}_{st}} \tilde{\pi} (\wp')\exp[-\theta \tilde{c}(\wp')]}
= \frac{\tilde{\pi}(\wp) \exp[-\theta \tilde{c}(\wp)]}{\mathcal{Z}}
\label{Eq_Boltzmann_probability_distribution01}
\end{equation}
where $\theta = 1/T$ is the inverse temperature and the denominator
\begin{equation}
\mathcal{Z} = \sum_{\wp\in\mathcal{P}_{st}} \\ \tilde{\pi} (\wp)\exp[-\theta \tilde{c}(\wp)]
\label{Eq_partition_function_definition01}
\end{equation}
is the \textbf{partition function} of the system. Eq.~\ref{Eq_Boltzmann_probability_distribution01} provides the randomized routing policy in terms of full paths from $s$ to $t$.

Interestingly,\footnote{This is also a standard result from statistical physics.} if we replace the probability distribution $\mathrm{P}(\cdot)$ by the optimal distribution $\mathrm{P}^{*}(\cdot)$ provided by Eq.\ \ref{Eq_Boltzmann_probability_distribution01} in the objective function (Eq.\ \ref{Eq_optimization_problem_BoP01}), we obtain
\begin{align}
\phi^{*} = \phi(\mathrm{P}^{*}) &= \dsum_{\wp \in \mathcal{P}_{st}} \mathrm{P}^{*}(\wp) \tilde{c}(\wp) + T \dsum_{\wp \in \mathcal{P}_{st}} \mathrm{P}^{*}(\wp) \log \left( \frac{\mathrm{P}^{*}(\wp)}{\tilde{\pi}(\wp)} \right) \nonumber \\
 &= \dsum_{\wp \in \mathcal{P}_{st}} \mathrm{P}^{*}(\wp) \tilde{c}(\wp) + T \dsum_{\wp \in \mathcal{P}_{st}} \mathrm{P}^{*}(\wp) \big( -\tfrac{1}{T} \tilde{c}(\wp) - \log \mathcal{Z} \big) \nonumber \\
 &= -T \log \mathcal{Z}
 \label{Eq_optimal_free_energy01}
\end{align}
which has been termed the directed \textbf{free energy distance} \cite{Kivimaki-2012}, and plays the role of a potential.

\subsection{Computing quantities of interest}

The quantities of interest can be computed by taking the partial derivative of the optimal free energy provided by Eq.\ \ref{Eq_optimal_free_energy01} \cite{Kivimaki-2012,Saerens-2008,Yen-08K}. Here, we introduce only the quantities that are necessary to deriving the algorithms developed later.

\paragraph{Fundamental matrix and partition function}

It turns out that the partition function can be computed in closed form from an auxiliary matrix,\footnote{Recall that the target node $t$ is made to be killing and absorbing by setting the corresponding row of the reference transition matrix to zero, which implies that row $t$ of $\mathbf{W}$ is also zero. Note also that the framework can easily be extended to any sub-stochastic matrix $\mathbf{W}$.} $\mathbf{W} = ( p_{ij}^{\mathrm{ref}} \exp[- \theta c_{ij}] )$. First, the \textbf{fundamental matrix} of the RSP system is defined as
\begin{equation}
\mathbf{Z} = \mathbf{I} + \mathbf{W} + \mathbf{W}^{2} + \cdots = (\mathbf{I} - \mathbf{W})^{-1} \quad \text{with} \quad \mathbf{W} = \mathbf{P}_{\mathrm{ref}} \circ \exp[-\theta \mathbf{C}]
\label{Eq_fundamentalMatrix01}
\end{equation}
where $\mathbf{C}$ is the cost matrix and $\circ$ is the elementwise (Hadamard) product. The equation sums up contributions of different paths lengths, starting from zero-length paths (identity matrix $\mathbf{I}$). The partition function is then provided by $\mathcal{Z} = [\mathbf{Z}]_{st} = z_{st}$  \cite{Kivimaki-2012,Saerens-2008,Yen-08K}.

\paragraph{Computation of flows and numbers of visits}

The directed \textbf{flow} in edge $(i,j) \in \mathcal{E}$ (the expected number of passages through $(i,j)$ when going from $s$ to $t$) can be obtained from Eq.\ \ref{Eq_optimal_free_energy01} and Eq.\ \ref{Eq_fundamentalMatrix01} for a given inverse temperature $\theta = 1/T$ (see \cite{Kivimaki-2016,Saerens-2008} for details) by
\begin{equation}
\bar{n}_{ij} \triangleq \dsum_{\wp\in\mathcal{P}_{st}} \mathrm{P}(\wp) \, \eta \big( (i,j) \in \wp \big) = - \tfrac{1}{\theta} \frac{\partial \log\mathcal{Z}} {\partial c_{ij}} = \frac{ z_{si} w_{ij} z_{jt} } {z_{st}}
\label{Eq_computation_edge_flows01}
\end{equation}
where $\eta \big( (i,j) \in \wp \big)$ counts the number of times edge $(i,j)$ appears on path $\wp$.
Now, as only the row $s$ and the column $t$ of fundamental matrix $\mathbf{Z}$ are needed, two systems of linear equations can be solved instead of the matrix inversion in Eq.\ \ref{Eq_fundamentalMatrix01}.
Note also that it can be readily shown from (\ref{Eq_computation_edge_flows01}) that flows are conserved on intermediate nodes $\mathcal{V} \setminus \{ s,t \}$; indeed, $\sum_{i=1}^{n} (\bar{n}_{ij} - \bar{n}_{ji}) = 0$ (input flow to $j$ is equal to output flow from $j$).
Let us further define the matrix containing the expected number of passages through edges $(i,j)$ by $\mathbf{N} = (\bar{n}_{ij})$.
From Eq.\ \ref{Eq_computation_edge_flows01}, the expected \textbf{number of visits} to a node $j$ can also be defined and easily computed as
\begin{equation}
\bar{n}_{j} = \sum_{i \in \mathcal{P}red(j)} \bar{n}_{ij} + \delta_{sj} = \sum_{i=1}^{n} \bar{n}_{ij} + \delta_{sj} = \frac{ z_{sj} z_{jt} } {z_{st}}
\label{Eq_computation_node_flows01}
\end{equation}
because a unit source flow of $+1$ is injected in node $s$. Note that $\mathcal{P}red(j)$ is the set of predecessor nodes of node $j$ and that we used $\sum_{i=1}^{n} z_{si} w_{ij} = z_{sj} - \delta_{sj}$, coming from $\mathbf{Z} (\mathbf{I} - \mathbf{W}) = \mathbf{I}$, with $\delta_{sj}$ being a Kronecker delta.

\paragraph{Optimal transition probabilities}

Finally, the optimal transition probabilities of following any edge $(i,j) \in \mathcal{E}$ (the policy) induced by the set of paths $\mathcal{P}_{st}$ and their probability mass (Eq.\ \ref{Eq_Boltzmann_probability_distribution01}) are \cite{Saerens-2008}
\begin{equation}
p^{*}_{ij} = \frac{\bar{n}_{ij}}{\bar{n}_{i}} = \frac{z_{jt}}{z_{it}} w_{ij}  \quad \text{for all } i \ne t
\label{Eq_biased_transition_probabilities01}
\end{equation}
and $p^{*}_{tj} = 0$ for the target node and all $j$. These transition probabilities define a \textbf{biased} \textbf{random walk} (an absorbing Markov chain) on $G$ -- the random walker is ``attracted" by the target node $t$. The lower the temperature, the larger the attraction. Interestingly, these transition probabilities do not depend on the source node, and they correspond to the optimal randomized strategy, or policy, thereby minimizing free energy (Eq.\ \ref{Eq_optimization_problem_BoP01}) for reaching the target node. Notice further that $\bar{n}_{ij} = \bar{n}_{i} p^{*}_{ij}$.

\section{The net flow randomized shortest paths dissimilarity}
\label{Sec_net_flow_RSP01}

In this section, we introduce the \textbf{net flow RSP} dissimilarity to extend the standard RSP dissimilarity developed in \cite{Kivimaki-2012,Saerens-2008,Yen-08K}. Similarly to the standard RSP, the \textbf{net flow RSP} corresponds to the expected cost for reaching target node $t$ from source node $s$, but with the important difference that \emph{net flows} are considered instead of raw flows. These measures are now introduced in this section.

\subsection{Definition of the net edge flow}

In some situations, such as electric networks \cite{Snell-1984}, only net flows matter. Intuitively, this means that the edge flows in opposite directions $i \rightarrow j$ and $j \rightarrow i$ compensate each other so that only the positive net flow, provided by $|\bar{n}_{ij} - \bar{n}_{ji}|$, is taken into account, where edge flows are given by Eq.\ \ref{Eq_computation_edge_flows01}.
In many situations, net flows look intuitively more natural because of the argument of flow compensation common to electricity.
Net flows have already been used in the RSP framework in order to define node betweenness measures \cite{Kivimaki-2016}, generalizing two previous models based on electric currents \cite{Brandes-2005b,Newman-05}. They are further investigated in this section in order to define a new dissimilarity measure between nodes of a graph.

Inspired by electric networks \cite{Snell-1984}, the non-negative net flow in each edge $(i,j)$, denoted here as $j_{ij}$, is defined from Eq.\ \ref{Eq_computation_edge_flows01} by
\begin{equation}
j_{ij} = \mathrm{max} \big( (\bar{n}_{ij} - \bar{n}_{ji}), 0 \big)
= \delta(\bar{n}_{ij} > \bar{n}_{ji}) \, (\bar{n}_{ij} - \bar{n}_{ji})
\label{Eq_net_flow_defined01}
\end{equation}
where $\delta(\mathrm{p}) = 1$ if proposition $\mathrm{p}$ is true and 0 otherwise. In matrix form,
\begin{equation}
\mathbf{J} = \mathrm{max} \big( (\mathbf{N} - \mathbf{N}^{\mathrm{T}}), \bigzero \big)
\label{Eq_net_flow_matrix_form01}
\end{equation}
where the maximum is taken elementwise. This means that, for each edge, the net flow is defined (that is, positive) in only one direction,\footnote{Another common convention is to consider $j_{ij} = (\bar{n}_{ij} - \bar{n}_{ji})$, and thus $j_{ji} = - j_{ij}$.} and is equal to zero in the other direction.
From their definition, net flows are also conserved on intermediate nodes $\mathcal{V} \setminus \{ s,t \}$, $\sum_{i=1}^{n} (j_{ij} - j_{ji}) = 0$ (net input flow to $j$ is equal to net output flow from $j$).
Interestingly, because the flow is equal to zero in one of the two edge directions, the net flow defines a directed graph from the source to the destination node, even if the original graph is undirected.

\subsection{Expected net cost and net RSP dissimilarity measure}

The \textbf{expected cost} until absorption by target node $t$ at temperature $T$ can easily be computed in closed form from the RSP formalism \cite{Saerens-2008}. This expected cost spread in the network has been used as a dissimilarity measure between nodes \cite{Kivimaki-2012,Yen-08K} and has been termed the directed \textbf{RSP dissimilarity}. More formally, the expected cost spread in the network is given by
\begin{equation}
\langle \tilde{c} \rangle= \dsum_{\wp \in \mathcal{P}_{st}} \mathrm{P}(\wp) \tilde{c}(\wp)
\label{Eq_definition_expected_cost01}
\end{equation}

Let us now express the cost along path $\wp$ as
$
\tilde{c}(\wp) = \sum_{(i,j) \in \mathcal{E}} \eta \big( (i,j) \in \wp \big) \, c_{ij} \nonumber
$, where we saw that $\eta \big( (i,j) \in \wp \big)$ is the number of times edge $(i,j)$ appears on path $\wp$ and $\mathcal{E}$ is the set of edges. Injecting this last result in Eq.\ \ref{Eq_definition_expected_cost01} provides
\begin{equation}
\langle \tilde{c} \rangle
= \sum_{(i,j) \in \mathcal{E}} \bigg( \underbracket[0.5pt][3pt]{ \dsum_{\wp \in \mathcal{P}_{st}} \mathrm{P}(\wp) \, \eta \big( (i,j) \in \wp \big) }_{\text{expected number of passages } \bar{n}_{ij}} \bigg) c_{ij}
= \sum_{(i,j) \in \mathcal{E}} \bar{n}_{ij} c_{ij}
\label{Eq_definition_expected_cost02}
\end{equation}
or, in matrix form,
\begin{equation}
\langle \tilde{c} \rangle = \mathbf{e}^{\mathrm{T}} (\mathbf{N} \circ \mathbf{C}) \mathbf{e}
\label{Eq_real_expected_cost01}
\end{equation}
where $\circ$ is the elementwise (Hadamard) matrix product, $\mathbf{e}$ is a column vector of 1s, and $\mathbf{N}$ is the matrix of expected number of passages through edges defined in Eq.\ \ref{Eq_computation_edge_flows01}. Intuitively, this quantity is just the cumulative sum of the expected number of passages through each edge times the cost of following the edge.

When dealing with net flows instead, Eq.~\ref{Eq_real_expected_cost01}, now computing the expected \emph{net cost}, becomes
\begin{equation}
\langle \tilde{c}_{\mathrm{net}} \rangle = \mathbf{e}^{\mathrm{T}} \big[ \max ( (\mathbf{N} - \mathbf{N}^{\mathrm{T}}), \bigzero ) \circ \mathbf{C} \big] \mathbf{e}
= \mathbf{e}^{\mathrm{T}} ( \mathbf{J} \circ \mathbf{C} ) \mathbf{e}
\label{Eq_real_expected_net_cost01}
\end{equation}
This quantity can be interpreted as the net cost needed to reach the target node $t$ from the source node $s$ in a biased random walk (defined by Eq.\ \ref{Eq_Boltzmann_probability_distribution01} or Eq.\ \ref{Eq_biased_transition_probabilities01}) attracting the walker toward the target node $t$. It is, therefore, the equivalent of the expected first passage cost defined in Markov chain theory, translated into the RSP formalism and for net flows. It can be seen as a directed dissimilarity between node $s$ and node $t$, taking both proximity and amount of connectivity between $s$ and $t$ into account.

When the temperature is low, $T \rightarrow 0^{+}$, the directed dissimilarity $\langle \tilde{c}_{\mathrm{net}} \rangle_{st}$ reduces to the least-cost dissimilarity between $s$ and $t$, while when $T \rightarrow \infty$, it tends to the expected net cost for a random walker moving according to the reference random walk (and thus electric current).
This quantity is in fact equivalent to the so-called $R_p$ distance introduced in \cite{Nguyen-2016} for $p = 1$, that is, the weighted-by-costs sum of the net flows in the case of a pure random walk (electric current).

Therefore, the \textbf{net flow RSP dissimilarity} (nRSP, the counterpart of the standard RSP dissimilarity  \cite{Kivimaki-2012,Saerens-2008,Yen-08K}) between node $s$ and node $t$ is defined as the symmetrized quantity
\begin{equation}
\myDelta^{\mathrm{n\textsc{rsp}}}_{st} = \langle \tilde{c}_{\mathrm{net}} \rangle_{st} + \langle \tilde{c}_{\mathrm{net}} \rangle_{ts}
\label{Eq_RSP_dissimilarity_definition01}
\end{equation}
where the starting and ending nodes are specified again. This is similar to the symmetric commute-cost quantity appearing in Markov chains \cite{FoussKDE-2005}, characterizing their relative accessibility \cite{Chebotarev-1997}.

Notice the difference between this quantity and the energy spread in an electric network. Indeed, if the costs are viewed as resistances, then, in the context of a resistive network, the energy weights the costs by the \emph{squared} net flow, instead of by the simple net flow in Eq.\ \ref{Eq_real_expected_net_cost01} \cite{Snell-1984}.

\subsection{Net flows define a directed acyclic graph}
\label{Subsec_electrical_current_DAG01}

Let us now show that the RSP net flows to a fixed target node $t$ define a directed acyclic graph (DAG) when the reference probabilities are defined by Eq.\ \ref{Eq_Transition_probabilities01} on a weighted undirected graph $G$. This comes from the fact that the net flows provided by Eq.\ \ref{Eq_net_flow_defined01} can be considered as an electric current generated from a new graph $\hat{G}_{t}$ derived from $G$ by redefining its edge weights. In addition, electric currents define a DAG because current always follows edges in the direction of decreasing potential (voltage). This potential therefore defines a topological ordering of the nodes, from a higher potential to a lower one (and lowest on the target node).

More precisely, for a fixed target $t$, let us define the graph $\hat{G}_{t}$ by considering the following weights on edges $(i,j)$ (conductances in electric circuits)
\begin{equation}
    \hat{a}_{ij} \triangleq z_{it} w_{ij} z_{jt} d_{i}
    = z_{it} p_{ij}^{\mathrm{ref}}  \exp[-\theta c_{ij}] z_{jt} d_{i}
    = z_{it} a_{ij}  \exp[-\theta c_{ij}] z_{jt}
\label{Eq_weights_new_graph01}    
\end{equation}
which is symmetric when $\mathbf{A}$ and $\mathbf{C}$ are symmetric (undirected graph). In this derivation, we used Eqs.\ \ref{Eq_Transition_probabilities01} and \ref{Eq_fundamentalMatrix01}.
In matrix form, this reads $\hat{\mathbf{A}} = \mathbf{Diag}( \mathbf{col}_{t}(\mathbf{Z}) ) (\mathbf{A} \circ \exp[\mathbf{-\theta C}]) \mathbf{Diag}( \mathbf{col}_{t}(\mathbf{Z}) )$ where $\mathbf{z}_{t} = \mathbf{col}_{t}(\mathbf{Z})$ extracts column $t$ (containing elements $z_{it}$) of matrix $\mathbf{Z}$ and $\mathbf{Diag}(\mathbf{z}_{t})$ defines a diagonal matrix from vector $\mathbf{z}_{t}$. 

The natural transition probabilities on this new graph $\hat{G}_{t}$ are provided by Eq.\ \ref{Eq_Transition_probabilities01} where we replace $a_{ij}$ by $\hat{a}_{ij}$,
\begin{equation}
\hat{p}_{ij} = \frac{\hat{a}_{ij}}{{\sum_{j'=1}^{n}} \hat{a}_{ij'}}
= \frac{w_{ij} z_{jt}}{\sum_{j'=1}^{n} w_{ij'} z_{j't}}
= \frac{z_{jt}}{z_{it}} w_{ij} \quad \text{for all } i \ne t
\label{Eq_biased_transition_probabilities_new_graph01}
\end{equation}
which, from Eq.\ \ref{Eq_biased_transition_probabilities01}, are exactly the optimal RSP transition probabilities. Note that we used the relation $z_{it} = \sum_{j'=1}^{n} w_{ij'} z_{j't} + \delta_{it}$,
which can be easily derived from the definition of the fundamental matrix, $(\mathbf{I} - \mathbf{W}) \mathbf{Z} = \mathbf{I}$ (Eq.\ \ref{Eq_fundamentalMatrix01}; see also, e.g., \cite{Kivimaki-2012,Saerens-2008,Yen-08K}).

This shows that the net flows resulting from the optimal biased random walk provided by Eq.\ \ref{Eq_biased_transition_probabilities01} are generated by a natural random walk on $\hat{G}_{t}$ where target node $t$ is made absorbing (an absorbing Markov chain). From the close relationship\footnote{Net flows defined by a random walk on an undirected graph $\hat{G}_{t}$, where node $t$ is made absorbing, correspond to the electric currents (see \cite{Snell-1984}, p. 50).} between random walks and electric current on an undirected graph \cite{Snell-1984}, this current defines a DAG on $\hat{G}_{t}$. From Eq.\ \ref{Eq_net_flow_defined01}, the corresponding \textbf{net flow transition probabilities} on the DAG are
\begin{equation}
p_{ij}^{\mathrm{net}*} = \frac{j_{ij}}{{\sum_{j'=1}^{n}} j_{ij'}}
\label{Eq_Transition_probabilities_net_flow01}
\end{equation}
Let us now turn to the description of an algorithm that enables all pairs of net flow distances to be computed on a graph.

\subsection{Computation of the net flow randomized shortest paths dissimilarity}

This subsection shows how the net flow RSP dissimilarity between all pairs of nodes (Eq.\ \ref{Eq_RSP_dissimilarity_definition01}) can be computed in matrix form on an undirected graph. Unfortunately, the computation of these net flow RSP dissimilarities is more time-consuming than computing the standard RSP dissimilarities, for which it suffices to perform a matrix inversion \cite{Kivimaki-2012}. This is because, before being able to compute the dissimilarities, we need to find the net flows, which involves a non-linear function (max). It is, however, still feasible for small- to medium-size networks.

The algorithm for computing the net flow RSP dissimilarity is detailed in Algorithm \ref{Alg_randomized_shortest_path_net_dissimilarity01}. It uses a trick introduced in \cite{Kivimaki-2016} for calculating the net flows between all source-destination pairs $s$, $t$ in a particular edge $(i,j)$ without having to explicitly turn node $t$ into a killing, absorbing node. More specifically, the procedure is a simple adaptation of Algorithm 2 in \cite{Kivimaki-2016} (following Eq.\ 12 in this work, providing net flows) to the case of an undirected graph and the computation of net flow dissimilarities, rather than betweenness centrality. It is also optimized in order to loop over (undirected) edges only once: on line \ref{Line_contribution}, the contributions of the two directions of edge $(i,j)$ (one is necessarily equal to $0$ and the other is equal to $| \bar{n}_{ij} - \bar{n}_{ji} |$) are summed together.

Following \cite{Kivimaki-2016}, its time complexity is $\mathcal{O}(n^3 + m n^2)$, where $m$ is the number of edges and $n$ the number of nodes, or $\mathcal{O}(m \, n^2)$ overall because $m \ge n$ for an undirected, connected, graph. Indeed, the algorithm contains a matrix inversion, which is $\mathcal{O}(n^3)$. Thereafter, there is a loop over all edges that repeats some standard matrix operations of order $\mathcal{O}(n^2)$, which finally provides $\mathcal{O}(n^3 + m n^2)$.
Therefore, the algorithm does not scale well on large graphs; in its present form, it can only be applied on medium-size graphs.

\begin{algorithm}[t!]
\caption[randomized shortest paths net flow dissimilarity]
{\small{Computing the net flow randomized shortest paths dissimilarity matrix (inspired by \cite{Kivimaki-2016}).}}

\algsetup{indent=2em, linenodelimiter=.}

\begin{algorithmic}[1]
\footnotesize
\REQUIRE $\,$ \\
 -- A weighted, undirected, connected graph $G$ containing $n$ nodes. \\
 -- The $n\times n$ symmetric adjacency matrix $\mathbf{A}$ associated to $G$, containing non-negative affinities.\\
 -- The $n\times n$ reference transition probabilities matrix $\mathbf{P}_{\mathrm{ref}}$ associated to $G$ (usually, the transition probabilities associated to the natural random walk on the graph, $\mathbf{P}_{\mathrm{ref}} = \mathbf{D}^{-1} \mathbf{A}$ where $\mathbf{D}$ is the outdegree diagonal matrix).\\
 -- The $n\times n$ symmetric cost matrix $\mathbf{C}$ associated to $G$, defining non-negative costs of transitions.\\
 -- The inverse temperature parameter $\theta > 0$.\\
 
\ENSURE $\,$ \\
 -- The $n \times n$ randomized shortest paths net flow dissimilarity matrix $\boldsymbol{\Delta}$ defined on all source-destination pairs.\\
\STATE $\mathbf{W} \leftarrow \mathbf{P}_{\mathrm{ref}}\circ\exp\left[-\theta\mathbf{C}\right]$ \COMMENT{elementwise exponential and multiplication $\circ$} \\
\STATE $\mathbf{Z} \leftarrow (\mathbf{I}-\mathbf{W}\mathbf{)}^{-1}$ \COMMENT{the fundamental matrix} \\
\STATE $\boldsymbol{\Delta} \leftarrow \bigzero$ \COMMENT{initialize the $n \times n$ net RSP flow dissimilarity matrix}
\FOR[compute contribution of each node $i$]{$i=1$ to $n$}
\STATE $\mat{z}^{\mathrm{c}}_{i} \leftarrow \mat{col}_{i}(\mat{Z})$, $\mat{z}^{\mathrm{r}}_{i} \leftarrow \mat{row}_{i}(\mat{Z})$ \COMMENT{copy column $i$ and row $i$ of $\mat{Z}$ transformed into a column vector}
\FOR[loop on neighboring nodes $j$, considering each (undirected) edge only once]{$j \in \mathcal{N}(i)$ with $j > i$}
\STATE $\mat{z}^{\mathrm{c}}_{j} \leftarrow \mat{col}_{j}(\mat{Z})$, $\mat{z}^{\mathrm{r}}_{j} \leftarrow \mat{row}_{j}(\mat{Z})$  \COMMENT{copy column $j$ and row $j$ of $\mat{Z}$ transformed into a column vector}
\STATE $\mat{N}_{ij}
\leftarrow w_{ij} \left[ \left( \mat{z}^{\mathrm{c}}_{i} (\mat{z}^{\mathrm{r}}_{j})^{\mathrm{T}} \div \mat{Z} \right)
- \mat{e} \left( (\mat{z}^{\mathrm{c}}_{i} \circ \mat{z}^{\mathrm{r}}_{j}) \div \mat{diag}(\mat{Z}) \right)^{\mathrm{T}} \right]$ \COMMENT{matrix of flow in edge $i \rightarrow j$ for all source-destination pairs (see \cite{Kivimaki-2016}, Eq.\ 17)}
\STATE $\mat{N}_{ji}
\leftarrow w_{ji} \left[ \left( \mat{z}^{\mathrm{c}}_{j} (\mat{z}^{\mathrm{r}}_{i})^{\mathrm{T}} \div \mat{Z} \right)
- \mat{e} \left( (\mat{z}^{\mathrm{c}}_{j} \circ \mat{z}^{\mathrm{r}}_{i}) \div \mat{diag}(\mat{Z}) \right)^{\mathrm{T}} \right]$ \COMMENT{matrix of flow in edge $j \rightarrow i$ for all source-destination pairs (see \cite{Kivimaki-2016}, Eq.\ 17)}
\STATE $\mathbf{Net}_{ij} \leftarrow \mathrm{abs} ( \mat{N}_{ij} - \mat{N}_{ji} )$ \COMMENT{net flow contribution from edge $i \leftrightarrow j$} \label{Line_contribution}
\STATE $\boldsymbol{\Delta} \leftarrow \boldsymbol{\Delta} + c_{ij} \mathbf{Net}_{ij} $ \COMMENT{update dissimilarity matrix with the contribution of edge $i \leftrightarrow j$} \\
\ENDFOR
\ENDFOR
\STATE $\boldsymbol{\Delta} \leftarrow \boldsymbol{\Delta} + \boldsymbol{\Delta}^{\mathrm{T}}$ \COMMENT{the resulting net RSP dissimilarities matrix}
\RETURN $\boldsymbol{\Delta}$

\end{algorithmic}
\label{Alg_randomized_shortest_path_net_dissimilarity01} 
\end{algorithm}

\section{Considering edge flow capacity constraints}
\label{Sec_edge_flow_capacity_constraints01}

In this section, an algorithm computing the optimal policy (the equivalent of Eq. \ref{Eq_biased_transition_probabilities01}) under flow capacity constraints on edges is derived. It is based on the fact that linear inequality constraints can easily be integrated in maximum entropy problems by considering Lagrangian duality (see, e.g., \cite{Jebara-2004} and references therein).

For convenience, we assume a weighted, undirected, connected graph $G$ with a single source node (node $s$) and a single target node (node $t \ne s$). An input flow is injected into node $s$ and absorbed by node $t$, but the model can easily be generalized for multiple sources and destinations. As before, it is assumed that the target node $t$ is killing and absorbing, meaning that the transition probabilities $p_{tj}^{\mathrm{ref}} = 0$ for all nodes $j$, including node $j = t$.

The idea now is to constrain the flow visiting some edges belonging to a set of constrained edges $\mathcal{C}$. The expected number of passages through these edges (see Eq.\ \ref{Eq_computation_edge_flows01}) is therefore forced not to exceed some predefined values (upper bound),
\begin{equation}
  \bar{n}_{ij} \le \sigma_{ij} \quad
\text{ for edges } (i,j) \in \mathcal{C}
\label{Eq_inequality_constraints_capacity01}
\end{equation}
which ensures that the flows on edges in $\mathcal{C}$ are limited by the capacities $\sigma_{ij} > 0$ and thus must remain in the interval $[0,\sigma_{ij}]$.
In this section, we consider that, although the graph $G$ is assumed undirected here, each capacity constraint is directed and thus active in only one direction of an edge. Therefore, each undirected edge $i \leftrightarrow j$ of $G$ possibly leads to two directed edges $(i,j)$ and $(j,i)$ in $\mathcal{C}$, reflecting a possibly different capacity constraint in each of the two directions. Thus, the set of constrained edges $\mathcal{C}$ contains directed edges, limiting the directional flow through them.

Moreover, we assume that the constraints are feasible, which is discussed later.\footnote{If not, the duality gap in our algorithm will not converge to zero.}
The problem aims to minimize the free energy objective function (Eq.\ \ref{Eq_optimization_problem_BoP01}) while satisfying these inequality constraints.

\subsection{The Lagrange function in case of capacity constraints}

From nonlinear optimization theory (see, e.g., \cite{Griva-2008}), the Lagrange function (following Eq. \ref{Eq_optimization_problem_BoP01}) is
\begin{align}
\mathscr{L}(\mathrm{P},\boldsymbol{\lambda})
&= \dsum_{\wp \in \mathcal{P}_{st}} \mathrm{P}(\wp) \tilde{c}(\wp) + T \dsum_{\wp \in \mathcal{P}_{st}} \mathrm{P}(\wp) \log \left( \frac{\mathrm{P}(\wp)}{\tilde{\pi}(\wp)} \right)
+ \mu \bigg( \dsum_{\wp \in \mathcal{P}_{st}} \mathrm{P}(\wp) - 1 \bigg) \nonumber \\
&\quad + \dsum_{(i,j) \in \mathcal{C}} \lambda_{ij} \big( \bar{n}_{ij} - \sigma_{ij} \big) \nonumber \\
&= \underbracket[0.5pt][3pt]{ \dsum_{\wp \in \mathcal{P}_{st}} \mathrm{P}(\wp) \tilde{c}(\wp) + T \dsum_{\wp \in \mathcal{P}_{st}} \mathrm{P}(\wp) \log \left( \frac{\mathrm{P}(\wp)}{\tilde{\pi}(\wp)} \right) }_{\text{free energy, }\phi(\mathrm{P})}
+ \mu \bigg( \dsum_{\wp \in \mathcal{P}_{st}} \mathrm{P}(\wp) - 1 \bigg) \nonumber \\
&\quad + \dsum_{(i,j) \in \mathcal{C}} \lambda_{ij} \bigg( \underbracket[0.5pt][3pt]{ \dsum_{\wp\in\mathcal{P}_{st}} \mathrm{P}(\wp) \, \eta \big( (i,j) \in \wp \big) }_{\bar{n}_{ij} \text{(see Eq.\ } \ref{Eq_computation_edge_flows01} )} - \sigma_{ij} \bigg) 
\label{Eq_Lagrange_node_flow_constraints_inequality01}
\end{align}
where there is a Lagrange parameter $\mu$ associated with the sum-to-one constraint and a Lagrange parameter $\lambda_{ij}$ associated with each constrained edge. The Lagrange parameters $\{ \lambda_{ij} \}$, $(i,j) \in \mathcal{C}$, are all non-negative in the case of inequality constraints \cite{Griva-2008} and are stacked into the parameter vector $\boldsymbol{\lambda}$.

Note that, by inspecting (\ref{Eq_Lagrange_node_flow_constraints_inequality01}) and from the convexity of the Kullback-Leibler divergence, the primal objective function to be minimized with respect to the discrete probabilities (which is similar to Eq.\ \ref{Eq_optimization_problem_BoP01}) is convex, the equality constraints are all linear, and the inequality constraints form a convex set. Therefore, the duality gap between the primal and dual problems is zero, which will be exploited for solving the problem \cite{Griva-2008}.

The Lagrange function in Eq.\ \ref{Eq_Lagrange_node_flow_constraints_inequality01} can be rearranged as
\begin{align}
\mathscr{L}(\mathrm{P},\boldsymbol{\lambda})
&= \dsum_{\wp \in \mathcal{P}_{st}} \mathrm{P}(\wp) \underbracket[0.5pt][3pt]{ \bigg( \tilde{c}(\wp) + \dsum_{(i,j) \in \mathcal{C}} \lambda_{ij} \, \eta\big( (i,j) \in \wp\big) \bigg) }_{ \text{augmented cost } \tilde{c}'(\wp) \text{ cumulated on path } \wp}
\nonumber \\
&\quad + T \dsum_{\wp \in \mathcal{P}_{st}} \mathrm{P}(\wp) \log \left( \frac{\mathrm{P}(\wp)}{\tilde{\pi}(\wp)} \right) + \mu \bigg( \dsum_{\wp \in \mathcal{P}_{st}} \mathrm{P}(\wp) - 1 \bigg)
- \dsum_{(i,j) \in \mathcal{C}} \lambda_{ij} \sigma_{ij} \nonumber \\
&= \dsum_{\wp \in \mathcal{P}_{st}} \mathrm{P}(\wp) \dsum_{(i,j) \in \mathcal{E}} \eta\big( (i,j) \in \wp \big) \underbracket[0.5pt][3pt]{ \big( c_{ij} + \delta\big((i,j) \in \mathcal{C}\big) \, \lambda_{ij} \big) }_{\text{augmented costs } c'_{ij}}
\nonumber \\
&\quad + T \dsum_{\wp \in \mathcal{P}_{st}} \mathrm{P}(\wp) \log \left( \frac{\mathrm{P}(\wp)}{\tilde{\pi}(\wp)} \right) + \mu \bigg( \dsum_{\wp \in \mathcal{P}_{st}} \mathrm{P}(\wp) - 1 \bigg)
- \dsum_{(i,j) \in \mathcal{C}} \lambda_{ij} \sigma_{ij} \nonumber \\
&= \underbracket[0.5pt][3pt]{ \dsum_{\wp \in \mathcal{P}_{st}} \mathrm{P}(\wp) \tilde{c}'(\wp) + T \dsum_{\wp \in \mathcal{P}_{st}} \mathrm{P}(\wp) \log \left( \frac{\mathrm{P}(\wp)}{\tilde{\pi}(\wp)} \right) }_{\text{free energy based on augmented costs, } \phi'(\mathrm{P})}
+ \mu \bigg( \dsum_{\wp \in \mathcal{P}_{st}} \mathrm{P}(\wp) - 1 \bigg) \nonumber \\
&\quad - \dsum_{(i,j) \in \mathcal{C}} \lambda_{ij} \sigma_{ij}
\label{Eq_lagrange_function_modified_transition01}
\end{align}
where the symbol $\delta\big((i,j) \in \mathcal{C}\big)$ is defined as $1$ when edge $(i,j) \in \mathcal{C}$ and $0$ otherwise. Note that we used $\tilde{c}(\wp) = \sum_{(i,j) \in \mathcal{E}} \eta\big( (i,j) \in \wp\big) \, c_{ij}$ (Eq.\ \ref{Eq_definition_expected_cost02}) to compute the total cost along path $\wp$.

During this derivation, we observed that the costs $c_{ij}$ can be redefined into \textbf{augmented costs} that integrate the additional ``virtual" costs (the Lagrange parameters) needed for satisfying the constraints, 
\begin{equation}
c'_{ij} =
\begin{cases}
c_{ij} + \lambda_{ij} &\text{when edge } (i,j) \in \mathcal{C} \\
c_{ij} &\text{otherwise}
\end{cases}
\label{Eq_redefined_costs_constrained_flow01}
\end{equation}
where $\mathbf{C}' = (c'_{ij})$ is the matrix containing these augmented costs. Thus, the Lagrange parameters have an interpretation similar to the dual variables in linear programming: they represent the extra cost to pay, associated with each edge, in order to satisfy the constraints \cite{Griva-2008}. This is also common in many network flow problems \cite{Ahuja-1993}.

Let $\phi'(\mathrm{P})$ be the free energy obtained in Eq.\ \ref{Eq_lagrange_function_modified_transition01} from these augmented costs (Eq.\ \ref{Eq_redefined_costs_constrained_flow01}). We now turn to the problem of finding the Lagrange parameters $\lambda_{ij}$ by exploiting Lagrangian duality.

\subsection{Exploiting Lagrangian duality}
\label{Subsec_Lagrangian_duality_edge_capacity_constraints01}

In this subsection, we will take advantage of the fact that, in the formulation of the problem (close to maximum entropy arguments), the Lagrange dual function and its gradient are relatively easy to compute; see, for instance, \cite{Jebara-2004} for similar arguments in the context of supervised classification.
Indeed, as the objective function is strictly convex, the equality constraints are linear and the support set for the path probabilities is convex, it is known that, provided that the problem is feasible,\footnote{Recall that we assume that the problem is feasible; see the discussion at the end of Subsection \ref{Subsec_algorithm_capacity_constraint01}.} there is only one global minimum and the duality gap is zero \cite{Griva-2008}. The optimum is a saddle point of the Lagrange function, and a common optimization procedure (sometimes called the Arrow-Hurwicz-Uzawa algorithm) consists of sequentially (i) solving the primal (finding the optimal probability distribution) while considering the Lagrange parameters as fixed, and then (ii) maximizing the dual (which is concave) with respect to the Lagrange parameters, until convergence. 

In our context, this provides the following steps \cite{Griva-2008} which are iterated,
\begin{equation}
\begin{cases}
\mathscr{L}(\boldsymbol{\lambda}) = \mathscr{L}(\mathrm{P}^{*}(\boldsymbol{\lambda}),\boldsymbol{\lambda})
= \minimize\limits_{\{ \mathrm{P}(\wp) \}_{\wp \in \mathcal{P}_{st}}} \mathscr{L}(\mathrm{P},\boldsymbol{\lambda}) &\text{\small{(compute the dual function)}} \\
\boldsymbol{\lambda}^{*} = \argmax\limits_{\boldsymbol{\lambda}} \mathscr{L}(\boldsymbol{\lambda}) &\text{\small{(maximize the dual function)}} \\
\boldsymbol{\lambda} = \boldsymbol{\lambda}^{*} &\text{\small{(update $\boldsymbol{\lambda}$)}}
\end{cases}
\label{Eq_primal_dual_lagrangian01}
\end{equation}
This is the procedure that will be followed whereby the dual function will be maximized through a simple gradient ascent procedure.

\subsubsection{Computing the dual function}

To compute the dual function $\mathscr{L}(\mathrm{P}^{*}(\boldsymbol{\lambda}),\boldsymbol{\lambda})$ in Eq.\ \ref{Eq_primal_dual_lagrangian01}, we first have to find the optimal probability distribution $\mathrm{P}^{*}$ in terms of the Lagrange parameters. We thus have to compute the minimum of Eq.\ \ref{Eq_lagrange_function_modified_transition01} for a constant $\boldsymbol{\lambda}$. But this Lagrange function (Eq.\ \ref{Eq_lagrange_function_modified_transition01}) is identical to the Lagrange function associated with the standard RSP optimization problem (Eq.\ \ref{Eq_optimization_problem_BoP01}), except that the costs $c_{ij}$ are replaced by the augmented costs $c'_{ij}$, and that the introduction of the final additional term does not depend on the probability distribution. Therefore, the probability distribution $\mathrm{P}(\cdot)$ minimizing Eq.\ \ref{Eq_lagrange_function_modified_transition01} is a Gibbs-Boltzmann distribution of the form of Eq.\ \ref{Eq_Boltzmann_probability_distribution01}, but it  now depends on the augmented costs instead of the original costs.

Next, we replace the probability distribution $\mathrm{P}(\cdot)$ in $\mathscr{L}(\mathrm{P},\boldsymbol{\lambda})$ by the optimal Gibbs-Boltzmann distribution $\mathrm{P}^{*}(\cdot)$, which depends on the augmented costs and thus also on the Lagrange parameters. From the result of Eq.\ \ref{Eq_optimal_free_energy01}, the obtained dual function\footnote{We leave out the sum-to-one constraint term which does not depend on $\boldsymbol{\lambda}$.} (Eq.\ \ref{Eq_primal_dual_lagrangian01}) is
\begin{equation}
\mathscr{L}(\boldsymbol{\lambda}) = \mathscr{L}(\mathrm{P}^{*}(\boldsymbol{\lambda}),\boldsymbol{\lambda}) = -T \log \mathcal{Z}' - \dsum_{(i,j) \in \mathcal{C}}  \lambda_{ij} \sigma_{ij}
\label{Eq_dual_lagrangian_flow_inequality_constraints02}
\end{equation}
In this equation, $\mathcal{Z}'$ is the partition function (see Eq.\ \ref{Eq_partition_function_definition01}), computed from the augmented costs $\mathbf{C}'$, which depends on $\boldsymbol{\lambda}$. We now need to maximize this dual function with respect to these Lagrange parameters.

\subsubsection{Maximizing the dual function}
\label{Eq_dual_function_maximization01}

The maximization of the dual function can be done, by, for example, using the simple method developed by Rockafellar (see \cite{Rockafellar-1973}, Eqs.\ 10 and 12).

But let us first compute the gradient of the dual function (Eq.\ \ref{Eq_dual_lagrangian_flow_inequality_constraints02}) with respect to the non-negative $\lambda_{ij}$ defined on edges $(i,j) \in \mathcal{C}$.
From $-T \, \partial \log\mathcal{Z} / \partial c_{ij} = \bar{n}_{ij}$ (Eq.\ \ref{Eq_computation_edge_flows01}) and $\partial c'_{ij} / \partial \lambda_{ij} = \delta\big( (i,j) \in \mathcal{C} \big)$ (Eq.\ \ref{Eq_redefined_costs_constrained_flow01}), this gradient is simply $\partial \mathscr{L}(\boldsymbol{\lambda}) / \partial \lambda_{ij} 
= \partial (-T \log \mathcal{Z}' - \sum_{(k,l) \in \mathcal{C}}  \lambda_{kl} \sigma_{kl}) /  \partial \lambda_{ij} = \bar{n}_{ij}  - \sigma_{ij}$, where $\bar{n}_{ij}$ is computed from the augmented costs.
It can be observed that we simply recover the expressions for the capacity constraints; this is actually a standard result when dealing with maximum entropy problems (see, e.g., \cite{Cover-2006,Jebara-2004}).

For computing the Lagrange parameters, we thus follow \cite{Rockafellar-1973} who proposed the following  (gradient-based) updating rule,\footnote{Note that \cite{Rockafellar-1973} uses $2 \alpha$ for the error correction term (instead of $\alpha$ here).}
\begin{equation}
\lambda_{ij} \leftarrow \max \big( \lambda_{ij} + \alpha (\bar{n}_{ij} - \sigma_{ij}), 0 \big)
\text{ for all } (i,j) \in \mathcal{C}
\label{Eq_lagrange_parameters_update01}
\end{equation}
which is guaranteed to converge in the concave case, as long as $\alpha$ is positive, is not too large, and the problem is feasible \cite{Rockafellar-1973}. This expression states that, if the flow in an edge $(i,j)$ is too large (that is, $\bar{n}_{ij} - \sigma_{ij} > 0$), the augmented cost should increase in order to reduce this flow. On the contrary, if the flow is below the capacity threshold, the augmented cost should decrease until eventually reach the normal cost value $c_{ij}$ (when $\lambda_{ij} = 0$). Notice also that, because costs are multiplied by $\theta$ in the model (see Eq.\ \ref{Eq_fundamentalMatrix01}), it becomes insensitive to cost variations when $\theta$ is small (close to $0$, an almost random walk behavior). We therefore set the $\alpha$ parameter proportional to $1 / \theta$ in Eq.\ \ref{Eq_lagrange_parameters_update01} during our experiments.

Of course, we could use other, more sophisticated and more efficient, optimization techniques (see, e.g., \cite{Griva-2008}), but this simple procedure worked satisfactorily for our tests on small- to medium-size graphs. The parameter $\alpha$ has to be tuned manually for each different dataset, but this was not a problem. However, in its present form, the algorithm does not scale to larger graphs.

\subsection{The resulting algorithm}
\label{Subsec_algorithm_capacity_constraint01}

The resulting algorithm is presented in Algorithm \ref{Alg_constrained_RSP01}.\footnote{In this pseudocode, $\mathbf{e}_{i}$ is a column (basis) vector of $0$s except on row $i$ where it contains a $1$.}
It computes the optimal policy (Eq.\ \ref{Eq_biased_transition_probabilities01}), minimizing the objective function (Eq.\ \ref{Eq_optimization_problem_BoP01}) while satisfying the inequality constraints (Eq.\ \ref{Eq_inequality_constraints_capacity01}).
This optimal policy guides the random walker to the target state with a trade-off between exploitation and exploration that is monitored by the inverse temperature parameter $\theta = 1/T$.
The different steps of the procedure are as follows:
\begin{itemize}
  \item Initialize the Lagrange parameters to $0$ and the augmented costs to the original edge costs $\mathbf{C}$.
  \item Iterate the following steps until convergence:
  \begin{itemize}
  \item The required elements of the fundamental matrix are recomputed from the current augmented costs (Eq.\ \ref{Eq_fundamentalMatrix01}) by solving two systems of linear equations.
  \item The expected number of passages through each edge (edge flows) is computed (Eq.\ \ref{Eq_computation_edge_flows01}).
  \item The Lagrange parameters and the augmented costs are updated (Eqs.\ \ref{Eq_lagrange_parameters_update01}, \ref{Eq_redefined_costs_constrained_flow01}).
  \end{itemize}
  \item Compute and return the optimal policy (transition probabilities) according to Eq.\ \ref{Eq_biased_transition_probabilities01}.
\end{itemize}
Because two systems of $n$ linear equations need to be solved at each iteration, its complexity is of the order $\mathcal{O}(k n^3)$, depending on the number of iterations $k$ needed for convergence. This number of iterations is unknown in advance but could become large depending on the problem and the gradient step. However, for sparse graphs, the complexity could be reduced by taking advantage of special numerical methods for solving sparse systems of linear equations.

Note also that lower bound (instead of upper bound) constraints on the expected flows have also been considered -- in this case the flow is constrained to be greater or equal (and not lesser or equal) to a threshold value. However, the resulting virtual costs (Lagrange parameters) in this case can become negative in order to augment the flow in the edge. This sometimes leads to numerical instabilities when some costs become negative and negative cycles appear. One quick fix is to prohibit negative costs; however, by doing this, the problem becomes unfeasible in some situations. Therefore, we decided to leave the study of lower-bounded capacities for further work.

As a last remark, note that it was assumed that the problem is feasible, which can be difficult to check in practice. Indeed, the model injects a unit flow into the network so that the maximum flow through the network must be at least equal to one. This means that we cannot blindly assign capacities because, in the case where the problem is not feasible, the algorithm does not converge. One way to check the overall capacity of the network is to run a standard max-flow algorithm \cite{Ahuja-1993,Korte-2018}.

\begin{algorithm}[t!]
\caption[Solving the randomized shortest paths problem on a graph problem with capacity constraints]
{\small{Randomized shortest paths with capacity constraints.}}

\algsetup{indent=2em, linenodelimiter=.}

\begin{algorithmic}[1]
\footnotesize
\REQUIRE $\,$ \\
 -- A weighted, undirected, connected graph $G$ containing $n$ nodes. Node $s$ is the source node and node $t$ the target node. \\
 -- The $n\times n$ reference transition probabilities matrix $\mathbf{P}_{\mathrm{ref}}$ associated to $G$. \\
 -- The $n\times n$ symmetric cost matrix $\mathbf{C}$ associated to $G$, defining non-negative costs of transitions. \\
 -- The set of constrained edges $\mathcal{C}$ (see the text for details). \\
 -- The set of non-negative capacities on flows, $\{\sigma_{ij}\}$, defined on the set of constrained edges, $(i,j) \in \mathcal{C}$. \\
 -- The inverse temperature parameter $\theta > 0$.\\
 -- The gradient ascent step $\alpha > 0$.\\
 
\ENSURE $\,$ \\
 -- The $n \times n$ randomized policy provided by the transition matrix $\mathbf{P}^{*}$, defining a biased random walk on $G$ satisfying the capacity constraints.\\

\STATE $\boldsymbol{\lambda} \leftarrow \mathbf{0}$ \COMMENT{initialize the $| \mathcal{C}| \times 1$ Lagrange parameters vector} \\
\STATE $\mathbf{C}' \leftarrow \mathbf{C}$ \COMMENT{initialize the augmented costs matrix} \\
\STATE Set row $t$ of matrix $\mathbf{P}_{\mathrm{ref}}$ to $\mathbf{0}^{\mathrm{T}}$ \COMMENT{target node $t$ is made absorbing and killing}
\REPEAT[main iteration loop]
\STATE $\mathbf{W} \leftarrow \mathbf{P}_{\mathrm{ref}} \, \circ \, \exp[-\theta\mathbf{C}']$ \COMMENT{update $\mathbf{W}$ matrix (elementwise exponential and multiplication $\circ$)} \\
\STATE Solve $(\mathbf{I}-\mathbf{W}) \mathbf{z}_{t} = \mathbf{e}_{t}$ \COMMENT{backward variables $ \mathbf{z}_{t}$ (column $t$ of the fundamental matrix $\mathbf{Z}$) with elements $z_{it}$} \\
\STATE Solve $(\mathbf{I}-\mathbf{W})^{\mathrm{T}} \mathbf{z}_{s} = \mathbf{e}_{s}$ \COMMENT{forward variables $ \mathbf{z}_{s}$ (row $s$ of the fundamental matrix $\mathbf{Z}$ viewed as a column vector) with elements $z_{si}$} \\

\STATE $\mathbf{N} \leftarrow \dfrac{ \mathbf{Diag}(\mathbf{z}_{s}) \mathbf{W} \mathbf{Diag}(\mathbf{z}_{t}) } { z_{st} } $ \COMMENT{compute the expected number of passages in each edge (see Eq. \ref{Eq_computation_edge_flows01})} \\

\FORALL[gradient ascent: update all quantities associated to constrained edges]{$(i,j) \in \mathcal{C}$}
\STATE $\lambda_{ij}  \leftarrow \max \big( \lambda_{ij} + \alpha (\bar{n}_{ij} - \sigma_{ij}), 0 \big) $ \COMMENT{update Lagrange parameters} \\
\STATE $c'_{ij} \leftarrow c_{ij} + \lambda_{ij}$ \COMMENT{update augmented costs} \\
\ENDFOR

\UNTIL{convergence}
\STATE $\mathbf{P}^{*} \leftarrow (\mathbf{Diag}(\mathbf{z}_{t}))^{-1} \mathbf{W} \mathbf{Diag}(\mathbf{z}_{t})$ \COMMENT{compute optimal policy}
\RETURN $\mathbf{P}^{*}$

\end{algorithmic} \label{Alg_constrained_RSP01}

\end{algorithm}

\section{Dealing with net flow capacity constraints}
\label{Sec_net_flow_capacity_constraints01}

Let us now consider the case where the capacities $\sigma_{ij} > 0$ with $(i,j) \in \mathcal{C}$ are defined on \emph{net flows} instead of raw flows. As before, it is assumed in this section that the original graph is undirected and that the adjacency matrix as well as the cost matrix are symmetric.
In this situation (see Eq.\ \ref{Eq_net_flow_defined01} and its discussion), we now consider that the constraints operate on the net flows instead of on the raw flows,
\begin{align}
  & j_{ij} = \mathrm{max} \big( (\bar{n}_{ij} - \bar{n}_{ji}), 0 \big) \le \sigma_{ij}, \nonumber \\
  &\text{or equivalently both}
  \begin{cases}
  \bar{n}_{ij} - \bar{n}_{ji} \le \sigma_{ij} \textnormal{ and} \\
  \bar{n}_{ji} - \bar{n}_{ij} \le \sigma_{ij}
  \end{cases}
  \text{for each edge } (i,j) \in \mathcal{C}
  \label{Eq_net_flow_constraints01}
\end{align}
In the net flows setting, we further assume that $\sigma_{ji}$ is always defined when there exists a capacity constraint $\sigma_{ij}$ and that $\sigma_{ji} = \sigma_{ij}$ (symmetry). In Eq.\ \ref{Eq_net_flow_constraints01}, only one among the two flow differences is positive so that the constraint only operates in this direction of the flow: the second constraint is automatically satisfied. This also means that only one of the two constraints can become active.

To summarize, if the set of constraint nodes $\mathcal{C}$ contains edge $(i,j)$, it also necessarily contains its reciprocal $(j,i)$ (they come as a pair) with the same capacity value, $\sigma_{ji} = \sigma_{ij}$. We now turn to the definition of the Lagrange function and the derivation of the  algorithm.

\subsection{The Lagrange function in case of net flow capacity constraints}

After rearranging the terms as in Eq.\ \ref{Eq_lagrange_function_modified_transition01} and, once again, using $\bar{n}_{ij} = \sum_{\wp\in\mathcal{P}_{st}} \mathrm{P}(\wp) \, \eta\big((i,j) \in \wp\big)$, the Lagrange function then becomes
\begin{align}
\mathscr{L}(\mathrm{P},\boldsymbol{\lambda})
&= \dsum_{\wp \in \mathcal{P}_{st}} \mathrm{P}(\wp) \tilde{c}(\wp) + T \dsum_{\wp \in \mathcal{P}_{st}} \mathrm{P}(\wp) \log \left( \frac{\mathrm{P}(\wp)}{\tilde{\pi}(\wp)} \right)
+ \mu \bigg( \dsum_{\wp \in \mathcal{P}_{st}} \mathrm{P}(\wp) - 1 \bigg) \nonumber \\
&\quad + \dsum_{(i,j) \in \mathcal{C}} \lambda_{ij} \big[ (\bar{n}_{ij} - \bar{n}_{ji}) - \sigma_{ij} \big] \nonumber \\
&= \dsum_{\wp \in \mathcal{P}_{st}} \mathrm{P}(\wp) \bigg( \tilde{c}(\wp) + \dsum_{(i,j) \in \mathcal{C}} \lambda_{ij} \, \big( \eta\big( (i,j) \in \wp\big) - \eta\big( (j,i) \in \wp\big) \big) \bigg)
\nonumber \\
&\quad + T \dsum_{\wp \in \mathcal{P}_{st}} \mathrm{P}(\wp) \log \left( \frac{\mathrm{P}(\wp)}{\tilde{\pi}(\wp)} \right) + \mu \bigg( \dsum_{\wp \in \mathcal{P}_{st}} \mathrm{P}(\wp) - 1 \bigg)
- \dsum_{(i,j) \in \mathcal{C}} \lambda_{ij} \sigma_{ij}
\label{Eq_Lagrange_node_net_flow_constraints_inequality01}
\end{align}

Now, from the symmetry of edges (edges are present in pairs; for each $(i,j)$ $\in$ $\mathcal{C}$: $(j,i)$ $\in$ $\mathcal{C}$ and $\sigma_{ij} = \sigma_{ji}$), we deduce $\sum_{(i,j) \in \mathcal{C}} \lambda_{ij} \, \eta\big( (j,i) \in \wp\big) = \sum_{(j,i) \in \mathcal{C}} \lambda_{ij} \eta\big( (j,i) \in \wp\big)  = \sum_{(i,j) \in \mathcal{C}} \lambda_{ji} \, \eta\big( (i,j) \in \wp\big)$. Injecting this result into Eq.\ \ref{Eq_Lagrange_node_net_flow_constraints_inequality01} and proceeding in the same way as in Eq.\ \ref{Eq_lagrange_function_modified_transition01} provides
\begin{align}
\mathscr{L}(\mathrm{P},\boldsymbol{\lambda})
&= \dsum_{\wp \in \mathcal{P}_{st}} \mathrm{P}(\wp) \underbracket[0.5pt][3pt]{ \bigg( \tilde{c}(\wp) + \dsum_{(i,j) \in \mathcal{C}} \big( \lambda_{ij} - \lambda_{ji} \big) \, \eta\big( (i,j) \in \wp\big) \bigg) }_{ \text{augmented cost } \tilde{c}'(\wp) \text{ cumulated on path } \wp}
\nonumber \\
&\quad + T \dsum_{\wp \in \mathcal{P}_{st}} \mathrm{P}(\wp) \log \left( \frac{\mathrm{P}(\wp)}{\tilde{\pi}(\wp)} \right) + \mu \bigg( \dsum_{\wp \in \mathcal{P}_{st}} \mathrm{P}(\wp) - 1 \bigg)
- \dsum_{(i,j) \in \mathcal{C}} \lambda_{ij} \sigma_{ij} \nonumber \\
&= \dsum_{\wp \in \mathcal{P}_{st}} \mathrm{P}(\wp) \dsum_{(i,j) \in \mathcal{E}} \eta\big( (i,j) \in \wp \big) \underbracket[0.5pt][3pt]{ \big( c_{ij} + \delta\big((i,j) \in \mathcal{C}\big) \, \big( \lambda_{ij} - \lambda_{ji} \big) \big) }_{\text{augmented costs } c'_{ij}}
\nonumber \\
&\quad + T \dsum_{\wp \in \mathcal{P}_{st}} \mathrm{P}(\wp) \log \left( \frac{\mathrm{P}(\wp)}{\tilde{\pi}(\wp)} \right) + \mu \bigg( \dsum_{\wp \in \mathcal{P}_{st}} \mathrm{P}(\wp) - 1 \bigg)
- \dsum_{(i,j) \in \mathcal{C}} \lambda_{ij} \sigma_{ij} \nonumber \\
&= \underbracket[0.5pt][3pt]{ \dsum_{\wp \in \mathcal{P}_{st}} \mathrm{P}(\wp) \tilde{c}'(\wp) + T \dsum_{\wp \in \mathcal{P}_{st}} \mathrm{P}(\wp) \log \left( \frac{\mathrm{P}(\wp)}{\tilde{\pi}(\wp)} \right) }_{\text{free energy based on augmented costs, } \phi'(\mathrm{P})}
+ \mu \bigg( \dsum_{\wp \in \mathcal{P}_{st}} \mathrm{P}(\wp) - 1 \bigg) \nonumber \\
&\quad - \dsum_{(i,j) \in \mathcal{C}} \lambda_{ij} \sigma_{ij}
\label{Eq_Lagrange_node_net_flow_constraints_inequality02}
\end{align}
which has exactly the same form as in the raw flow case (see Eq.\ \ref{Eq_lagrange_function_modified_transition01}) with the exception that the definition of the augmented costs differs in the two expressions.
Indeed, as before, the costs $c_{ij}$ can be redefined into \emph{augmented} costs,
\begin{equation}
c'_{ij} =
\begin{cases}
c_{ij} + \lambda_{ij} - \lambda_{ji}  &\text{when edge } (i,j) \in \mathcal{C} \\
c_{ij} &\text{otherwise}
\end{cases}
\label{Eq_redefined_costs_constrained_net_flow01}
\end{equation}
We must stress the requirement that the constraints in Eq.\ \ref{Eq_net_flow_constraints01} be symmetric and come by pair. In addition, as discussed before, only one constraint can become active among the two directions $(i,j)$ and $(j,i)$, which implies that one of the two Lagrange multipliers $\{ \lambda_{ij}, \lambda_{ji} \}$ must be equal to $0$: the $\lambda_{ij}$ for which $(\bar{n}_{ij} - \bar{n}_{ji}) < 0$.

Moreover, by following the same reasoning as in the previous section (see Eqs. \ref{Eq_dual_lagrangian_flow_inequality_constraints02} and \ref{Eq_lagrange_parameters_update01}), it can immediately be observed that the dual function has the same form as before and is provided by Eq.\ \ref{Eq_dual_lagrangian_flow_inequality_constraints02}.

\subsection{The resulting algorithm}
\label{Subsec_resulting_algorithm_net_flows01}

In the net flow context, by proceeding in the same way as in the previous section (see Subsection \ref{Eq_dual_function_maximization01}), the gradient of the dual Lagrange function (Eq.\ \ref{Eq_dual_lagrangian_flow_inequality_constraints02}) for augmented costs provided by Eq.\ \ref{Eq_redefined_costs_constrained_net_flow01} is $d \mathscr{L}(\boldsymbol{\lambda}) / d \lambda_{ij} = ( \partial \mathscr{L}(\boldsymbol{\lambda}) / \partial c'_{ij} ) (\partial c'_{ij} / \partial \lambda_{ij}) + ( \partial \mathscr{L}(\boldsymbol{\lambda}) / \partial c'_{ji} ) (\partial c'_{ji} / \partial \lambda_{ij}) + \partial \mathscr{L}(\boldsymbol{\lambda}) / \partial \lambda_{ij} = \bar{n}_{ij} - \bar{n}_{ji} - \sigma_{ij}$.
From this last result, we can derive the update of the Lagrange parameters $\lambda_{ij}$ with $(i,j) \in \mathcal{C}$,

\begin{equation}
\lambda_{ij} \leftarrow
\begin{cases}
\max \big( \lambda_{ij} + \alpha (\bar{n}_{ij} - \bar{n}_{ji} - \sigma_{ij}), 0 \big)
&\text{when } (\bar{n}_{ij} - \bar{n}_{ji}) \ge 0  \\
0 &\text{when } (\bar{n}_{ij} - \bar{n}_{ji}) < 0
\end{cases}
\label{Eq_lagrange_parameters_net_flow_uptate01}
\end{equation}

Algorithm \ref{Alg_constrained_RSP01} is easy to adapt in order to consider net flow capacity constraints (with $\sigma_{ij} = \sigma_{ji}$); only lines 10 and 11 must be modified according to Eqs.\ \ref{Eq_redefined_costs_constrained_net_flow01} and \ref{Eq_lagrange_parameters_net_flow_uptate01}.

Notice that still another procedure could be derived when considering the edges as undirected and unique; that is, there is a unique affinity and cost value associated to each edge $i \leftrightarrow j$. In that situation, the Lagrange function is defined in terms of $m$ costs (instead of $2m$ in this work) and augmented costs cannot differ along the edge direction. A similar algorithm updating only one Lagrange parameter per edge could be derived for this case.

In practice, we however observed that the net flow constraint algorithm is slower to converge and more sensitive to the gradient step than simple flow constraints; in other words, the problem looks harder to solve. This is especially the case for small values of $\theta$, where the process behaves more like a random walker. Indeed, in that situation, the addition of capacity constraints seems to be partly in conflict with the objective of walking randomly and moving back and forth. One way to handle this issue \cite{Dial71} would be first to define a DAG (deduced, e.g., from the electric potential on nodes when imposing a +1 potential at the source node and a 0 potential at the target node, followed by a topological sort). This will enable an RSP problem with simple capacity constraints to be solved on this DAG. This has the additional advantage that it should be more efficient (we avoid cycles and can use dynamic programming techniques to compute the RSP solution) and scale to larger graphs. This extension is left for further work.

\section{Experiments}
\label{Sec_experiments01}

In this section, we first present two illustrative examples of the use of capacity constraints on the edge flows of a graph as well as a brief study of the scalability of the net flow Algorithm \ref{Alg_randomized_shortest_path_net_dissimilarity01}. Following this, we evaluate the net flow RSP on unsupervised classification tasks and compare its results to other state-of-the-art graph node distances. It is important to emphasize that our goal here was not to propose that new node clustering algorithms outperform state-of-the-art techniques. Rather, the aim was to investigate if the net flow RSP model is able to capture the community structure of networks in an accurate way, compared to other dissimilarity measures between nodes.

\subsection{Illustrative examples}

\begin{figure}[t]
\centering  
\subfigure[Expected number of visits to nodes (see Eq.\ \ref{Eq_computation_edge_flows01}), with $\theta=0.05$ and $\alpha=40$. Red represents a higher expected number of visits and blue a lower number of visits.]{\includegraphics[width=0.30\linewidth]{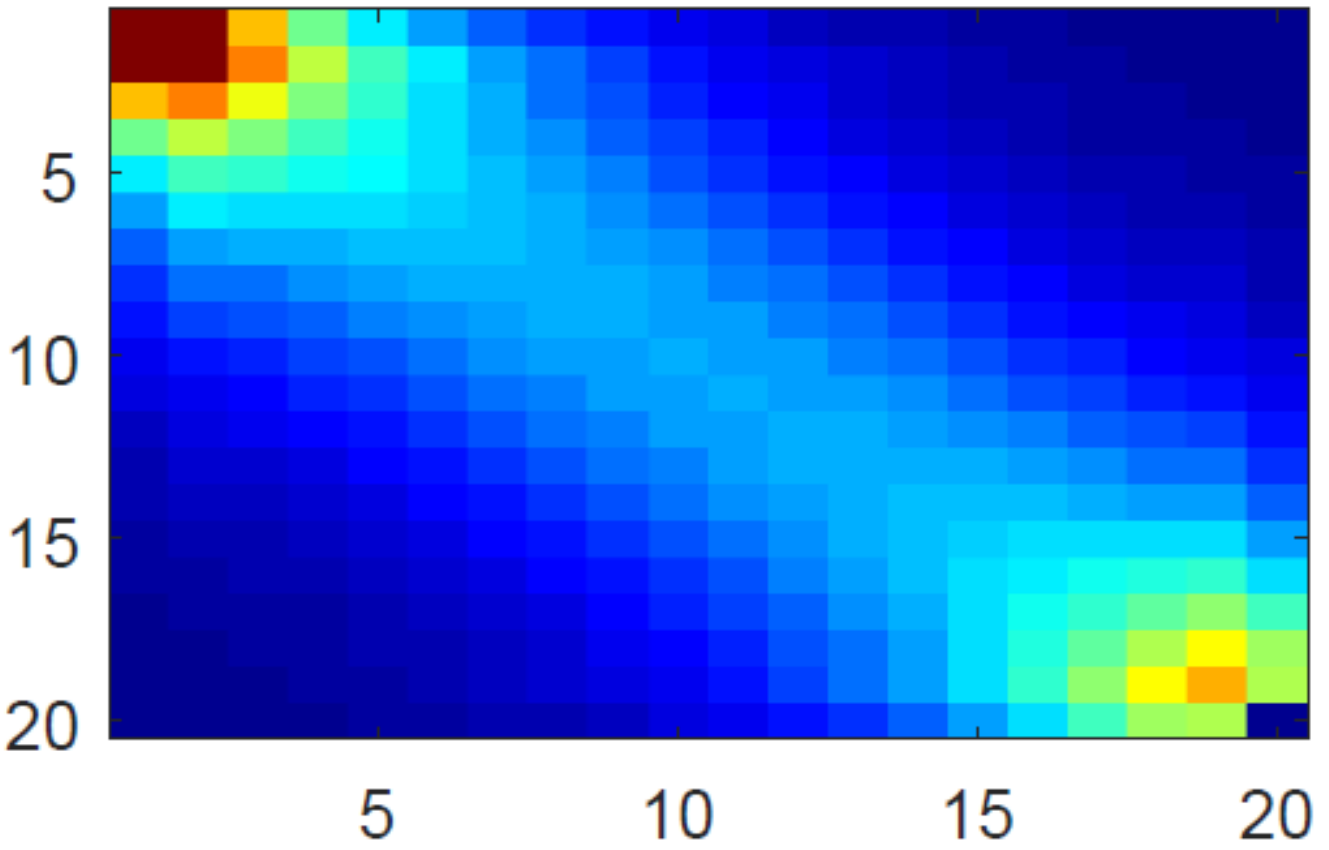}
\label{fig:PassageA}}
\hspace{2mm}
\subfigure[Introduction of capacity constraints on edges belonging to the obstacles (walls in red). Source node $s$ (upper left corner) and target node $t$ (lower right corner) are in green.]{\includegraphics[width=0.30\linewidth]{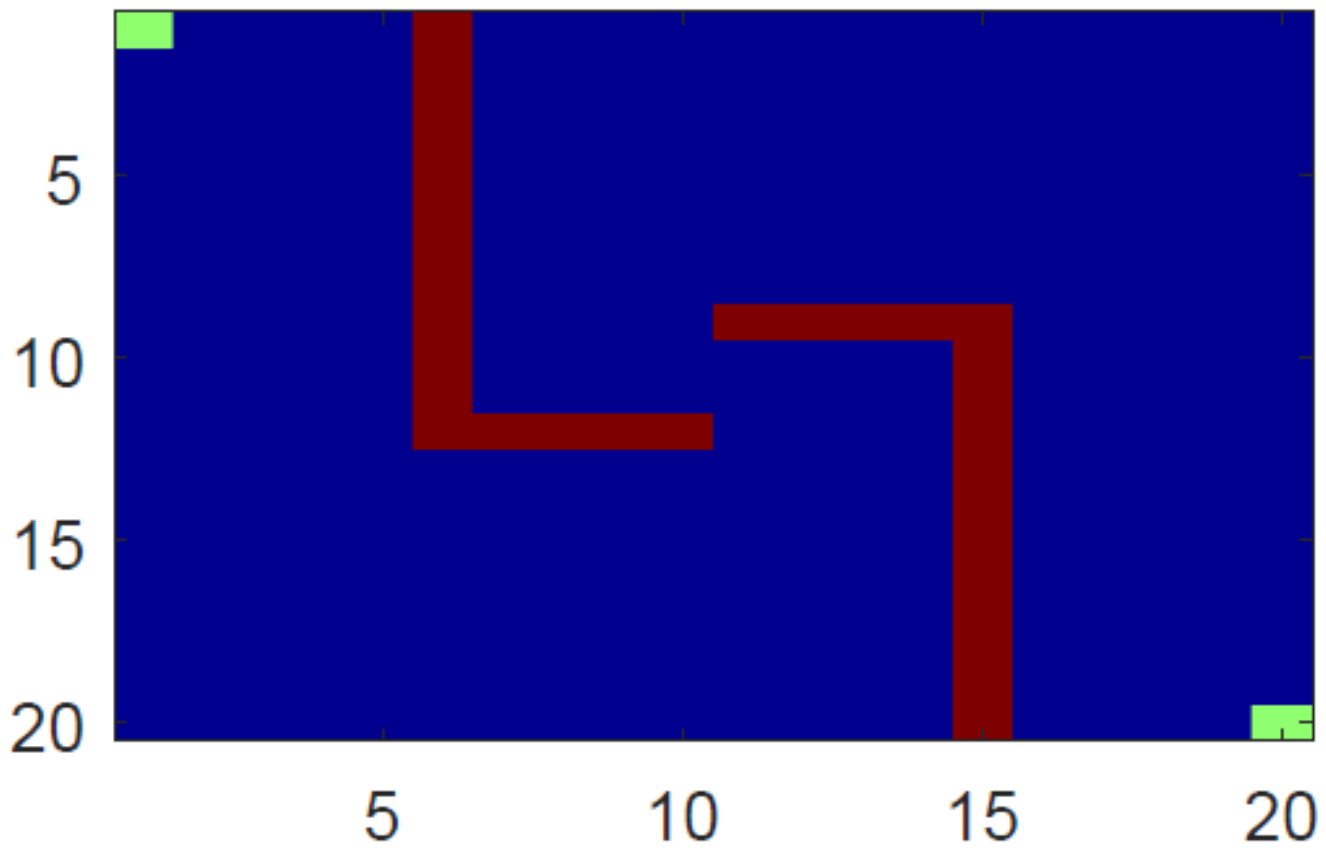}
\label{fig:PassageB}}
\hspace{2mm}
\subfigure[Expected number of visits to nodes (see Eq.\ \ref{Eq_computation_edge_flows01}) after introducing capacity constraints on the walls, for an intermediate value of $\theta=0.05$ and gradient step $\alpha=40$. 
]{\includegraphics[width=0.30\linewidth]{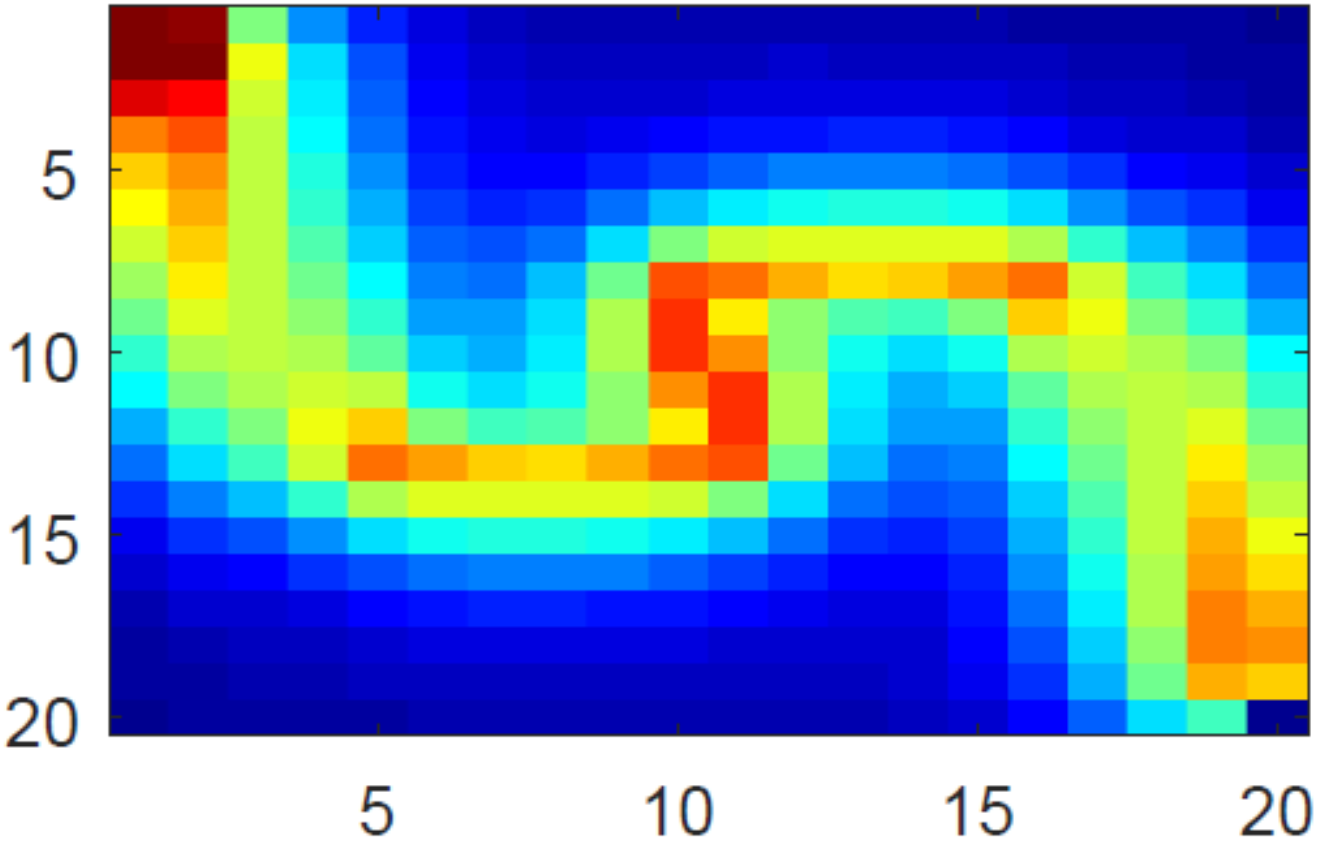}
\label{fig:PassageC}}
\caption{Illustrative example of the capacity-constrained RSP on a 2D grid.}
\label{fig:Passage}
\end{figure}

\paragraph{First example}
The first example illustrates the expected number of visits to nodes (see Eq.\ \ref{Eq_computation_node_flows01}) over the RSP paths distribution between one source node $s$ and one target node $t$ on a $20 \times 20$ grid, in two different situations obtained after running Algorithm \ref{Alg_constrained_RSP01}. Nodes were linked to their neighbors\footnote{Only horizontal and vertical neighbors are considered (no diagonal edge).} with a unit affinity and a unit cost. In the first situation, we did not set any capacity constraint, and the expected number of visits is represented in Fig.\ \ref{fig:PassageA}. As expected, walkers followed a trajectory close to the diagonal of the grid, representing the shortest paths between the source node and the target node.

In the second situation, we placed two obstacles (porous walls) by constraining the capacities of all the edges linking the nodes represented in red in Fig.\  \ref{fig:PassageB} to $0.01$. As can be observed in \ref{fig:PassageC}, in this situation, the expected number of visits to nodes no longer concentrated around the diagonal, but instead closely followed the obstacles. This trajectory reflects the least-cost paths between the source node and the target node, avoiding the low-capacity obstacles.

\begin{figure}[t]
     \begin{minipage}[t]{.45\linewidth}
        \scriptsize
\centering
\begin{tikzpicture}[shorten >=1pt, auto, node distance=1cm, ultra thick,
   classic/.style={circle,draw=black,font=\scriptsize\bfseries},
   edge_style/.style={draw=black},scale=0.8,transform shape]
   
    \node[classic] (S) at (1,0) {S};
    \node[classic] (a) at (3,2) {a};
    \node[classic] (b) at (3,0) {b};
    \node[classic] (c) at (3,-2) {c};
    \node[classic] (d) at (6,1) {d};
    \node[classic] (e) at (6,-2) {e};
    \node[classic] (f) at (8,2) {f};
    \node[classic] (g) at (8,0) {g};
    \node[classic] (h) at (8,-2) {h};
    \node[classic] (T) at (10,0) {T};
    
    \node[fill=white] (Inf) at (5.5,3.25) {$\infty$};
    
    \draw[edge_style]  (S) edge node{4} (a);
    \draw[edge_style]  (S) edge node{7} (b);
    \draw[edge_style]  (S) edge node{7} (c);
    \draw[edge_style]  (a) edge node{5} (b);
    \draw[edge_style]  (a) edge node{2} (d);
    \draw[edge_style]  (b) edge node{6} (d);
    \draw[edge_style]  (b) edge node{3} (e);
    \draw[edge_style]  (c) edge node{1} (e);
    \draw[edge_style]  (d) edge node{4} (f);
    \draw[edge_style]  (d) edge node{7} (g);
    \draw[edge_style]  (d) edge node{4} (h);
    \draw[edge_style]  (e) edge node{10} (h);
    \draw[edge_style]  (f) edge node{2} (g);
    \draw[edge_style]  (f) edge node{1} (T);
    \draw[edge_style]  (g) edge node{10} (T);
    \draw[edge_style]  (h) edge node{11} (T);
    
    \draw[edge_style,dotted,->] (S) |- (1,3) |- (10,3) -- (T);
 
\end{tikzpicture}
\caption{A small undirected graph composed of eight nodes \cite{Price-1971}. The values on the edges represent capacities; moreover, all edge costs are set to $1$, except the added shortcut edge (dashed line) whose cost is $10$. The maximum possible flow between the source node $s$ and the target node $t$ is 12, and is equal to the min-cut between nodes $\{ a,b,c \}$ and $\{ d,e \}$.}
\label{fig:GraphUndirected}
    \end{minipage}
    \hfill
    \begin{minipage}[t]{.45\linewidth}
        \centering
        \includegraphics[width=6.5cm]{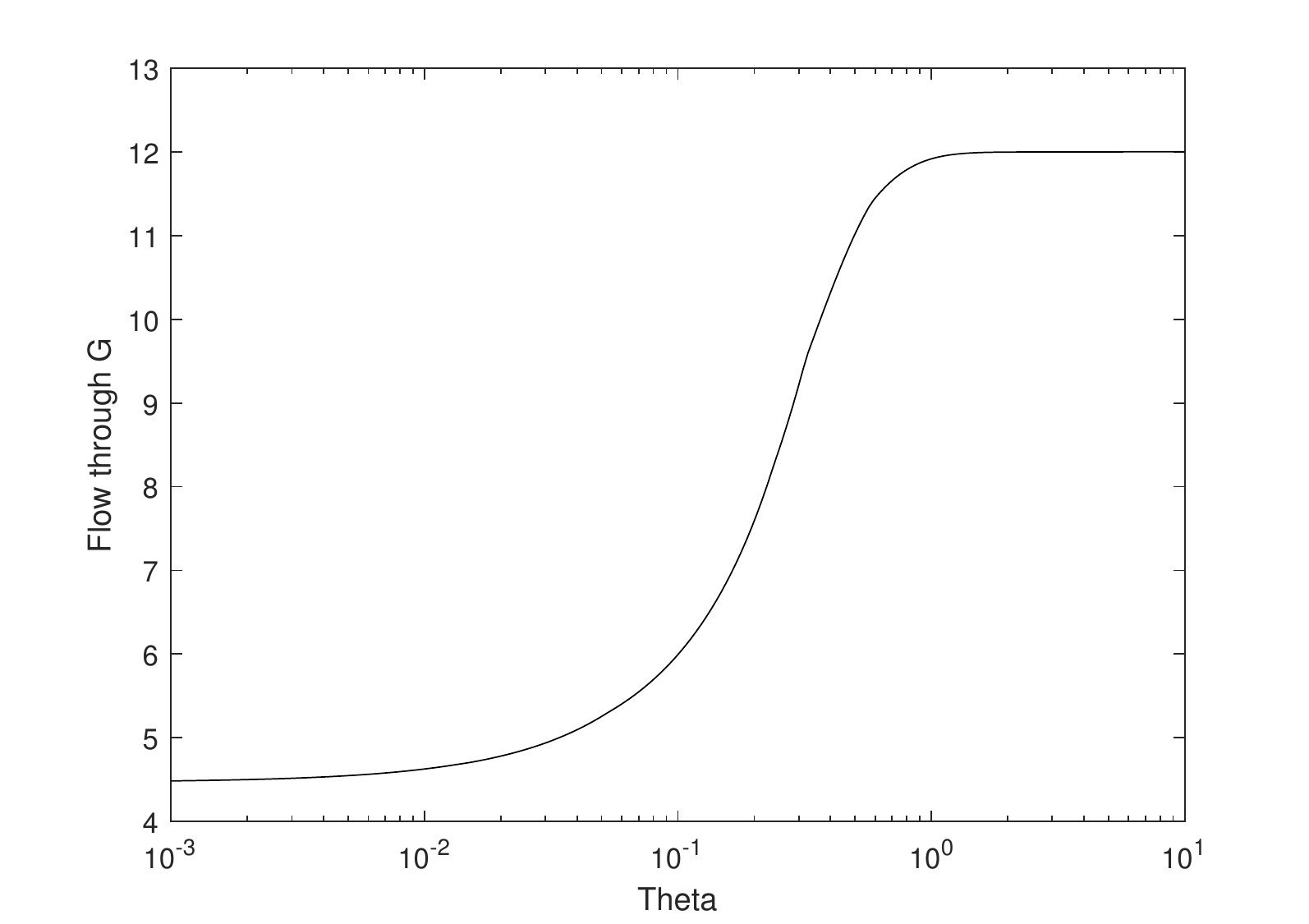}
        \caption{Evolution of the flow between nodes $\{ f,g,h \}$ and $t$ satisfying the capacity constraints (total flow in the original graph $G$, without the shortcut edge), provided by Algorithm \ref{Alg_constrained_RSP01}, in terms of the $\theta$ parameter, with a gradient step $\alpha = 1 / \theta$. The intersection between the curve and the $y$-axis when $\theta \rightarrow 0^{+}$ is 4.699. Note that the $x$-axis is scaled logarithmically.}
        \label{fig:EvMaxflow}        
     \end{minipage}
 \end{figure}

\paragraph{Second example}
Our second illustrative example was taken from \cite{Price-1971} and makes a link between the RSP with net flow capacity constraints and the maximum flow problem. Its aim was to show that Algorithm \ref{Alg_constrained_RSP01} could be used to compute the maximum flow (which takes a value of 12 in this example) between the source node $s$ and the target node $t$ in the undirected graph $G$ presented in Fig.\  \ref{fig:GraphUndirected}. Each edge of this graph has a unit cost and affinity, as well as capacities shown on the drawing. Note that, in practice, because the RSP model assumes a unit input flow, all capacities are scaled\footnote{We divide capacities by a graph cut value (an upper bound on the min-cut), such as the cut between nodes $\{ f,g,h \}$ and $t$: 22 in our example.} so that the value of the computed maximum flow after scaling lies between $0$ and $1$. The maximum flow of $G$ is then obtained by the reverse transformation.

To find the max-flow, we added a directed edge from $s$ to $t$ (dashed line) with infinite capacity and an edge cost of 10 to the original graph. This new edge introduced a ``shortcut" allowing the passage of all the overflow of the graph, but with a higher cost. Theoretically, our algorithm will avoid going through this shortcut as much as possible because it has a high cost compared to the other trajectories in the graph. Therefore, it should try to maximize the flow that travels through the original graph before using this shortcut edge.

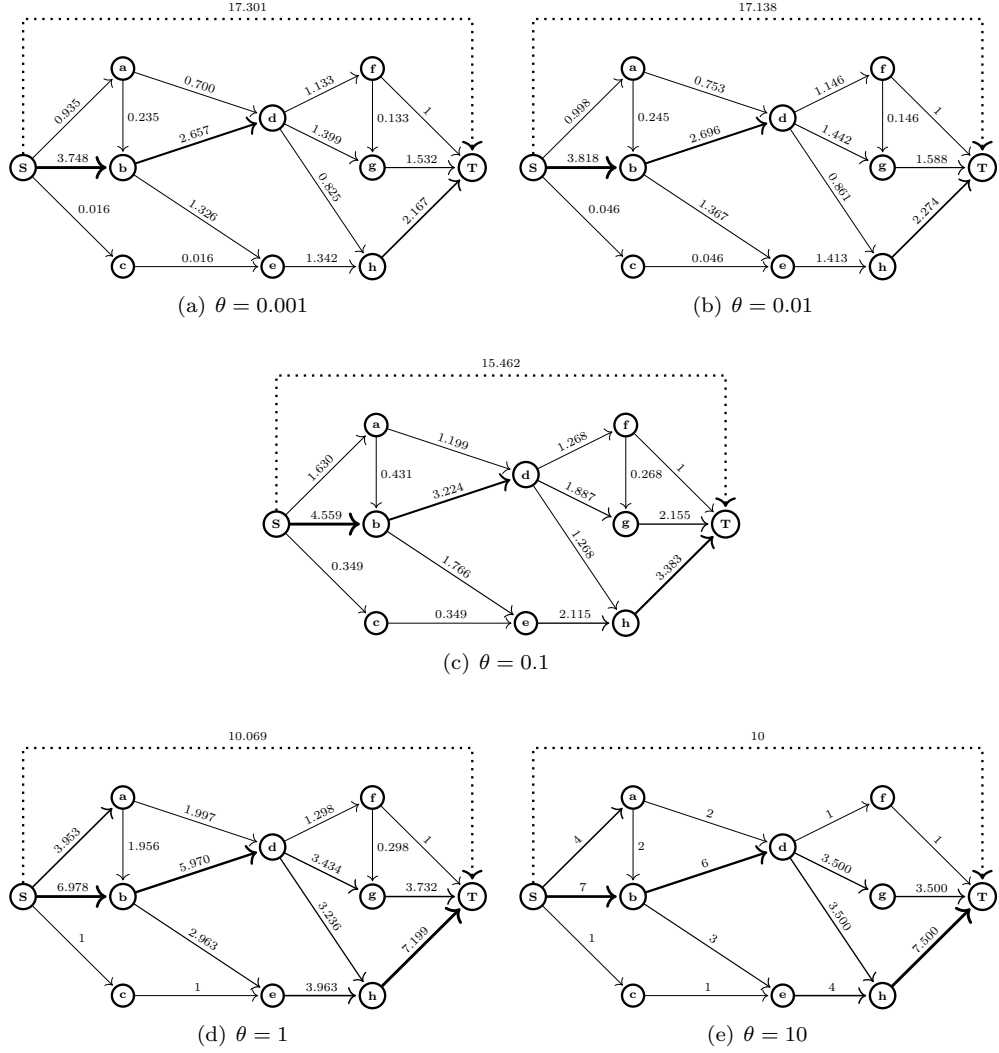
\begin{figure}[t!]
\subfigure[$\theta=0.001$]{
\resizebox{.49\linewidth}{!}{
        \scriptsize
\centering
\begin{tikzpicture}[shorten >=1pt, auto, node distance=1cm, thick,
   classic/.style={circle,draw=black,font=\scriptsize\bfseries},
   edge_style/.style={draw=black},scale=0.8,transform shape]
   
       \node[classic] (S) at (1,0) {S};
    \node[classic] (a) at (3,2) {a};
    \node[classic] (b) at (3,0) {b};
    \node[classic] (c) at (3,-2) {c};
    \node[classic] (d) at (6,1) {d};
    \node[classic] (e) at (6,-2) {e};
    \node[classic] (f) at (8,2) {f};
    \node[classic] (g) at (8,0) {g};
    \node[classic] (h) at (8,-2) {h};
    \node[classic] (T) at (10,0) {T};
    
    \node[fill=white] (Inf) at (5.5,3.25) {17.301};
    
    \draw[edge_style,->,line width=0.25pt]  (S) edge node[sloped]{0.935} (a); 
    \draw[edge_style,->,line width=1pt]  (S) edge node{3.748} (b); 
    \draw[edge_style,->,line width=0.01pt] (S) edge node{0.016} (c); 
    \draw[edge_style,->,line width=0.06pt] (a) edge node{0.235} (b);
    \draw[edge_style,->,line width=0.19pt]  (a) edge node[sloped]{0.700} (d); 
    \draw[edge_style,->,line width=0.71pt]  (b) edge node[sloped]{2.657} (d);
    \draw[edge_style,->,line width=0.35pt] (b) edge node[sloped]{1.326} (e);
    \draw[edge_style,->,line width=0.01pt] (c) edge node[sloped]{0.016} (e);
    \draw[edge_style,->,line width=0.30pt]  (d) edge node[sloped]{1.133} (f);
    \draw[edge_style,->,line width=0.37pt]  (d) edge node[sloped]{1.399} (g);
    \draw[edge_style,->,line width=0.22pt] (d) edge node[sloped]{0.825} (h);
    \draw[edge_style,->,line width=0.36pt]  (e) edge node{1.342} (h);
    \draw[edge_style,->,line width=0.04pt]  (f) edge node{0.133} (g);
    \draw[edge_style,->,line width=0.27pt] (f) edge node[sloped]{1} (T);
    \draw[edge_style,->,line width=0.41pt]  (g) edge node{1.532} (T);
    \draw[edge_style,->,line width=0.58pt]  (h) edge node[sloped]{2.167} (T);
   \draw[edge_style,dotted,->] (S) |- (1,3) |- (10,3) -- (T);
\end{tikzpicture}
    }}
    \hfill
    \subfigure[$\theta=0.01$]{
\resizebox{.49\linewidth}{!}{
        \scriptsize
\centering
\begin{tikzpicture}[shorten >=1pt, auto, node distance=1cm, thick,
   classic/.style={circle,draw=black,font=\scriptsize\bfseries},
   edge_style/.style={draw=black},scale=0.8,transform shape]
   
       \node[classic] (S) at (1,0) {S};
    \node[classic] (a) at (3,2) {a};
    \node[classic] (b) at (3,0) {b};
    \node[classic] (c) at (3,-2) {c};
    \node[classic] (d) at (6,1) {d};
    \node[classic] (e) at (6,-2) {e};
    \node[classic] (f) at (8,2) {f};
    \node[classic] (g) at (8,0) {g};
    \node[classic] (h) at (8,-2) {h};
    \node[classic] (T) at (10,0) {T};
    
    \node[fill=white] (Inf) at (5.5,3.25) {17.138};
    
    \draw[edge_style,->,line width=0.26pt]  (S) edge node[sloped]{0.998} (a);
    \draw[edge_style,->,line width=1pt]  (S) edge node{3.818} (b);
    \draw[edge_style,->,line width=0.01pt] (S) edge node{0.046} (c);
    \draw[edge_style,->,line width=0.06pt] (a) edge node{0.245} (b);
    \draw[edge_style,->,line width=0.20pt]  (a) edge node[sloped]{0.753} (d);
    \draw[edge_style,->,line width=0.71pt]  (b) edge node[sloped]{2.696} (d);
    \draw[edge_style,->,line width=0.36pt] (b) edge node[sloped]{1.367} (e);
    \draw[edge_style,->,line width=0.01pt] (c) edge node[sloped]{0.046} (e);
    \draw[edge_style,->,line width=0.30pt]  (d) edge node[sloped]{1.146} (f);
    \draw[edge_style,->,line width=0.38pt]  (d) edge node[sloped]{1.442} (g);
    \draw[edge_style,->,line width=0.23pt] (d) edge node[sloped]{0.861} (h);
    \draw[edge_style,->,line width=0.37pt]  (e) edge node{1.413} (h);
    \draw[edge_style,->,line width=0.04pt]  (f) edge node{0.146} (g);
    \draw[edge_style,->,line width=0.26pt] (f) edge node[sloped]{1} (T);
    \draw[edge_style,->,line width=0.42pt]  (g) edge node{1.588} (T);
    \draw[edge_style,->,line width=0.60pt]  (h) edge node[sloped]{2.274} (T);
   \draw[edge_style,dotted,->] (S) |- (1,3) |- (10,3) -- (T);
\end{tikzpicture}
    }}
    \begin{center}
\subfigure[$\theta=0.1$]{
\resizebox{.49\linewidth}{!}{
     
        \scriptsize
\centering
\begin{tikzpicture}[shorten >=1pt, auto, node distance=1cm, thick,
   classic/.style={circle,draw=black,font=\scriptsize\bfseries},
   edge_style/.style={draw=black},scale=0.8,transform shape]
   
       \node[classic] (S) at (1,0) {S};
    \node[classic] (a) at (3,2) {a};
    \node[classic] (b) at (3,0) {b};
    \node[classic] (c) at (3,-2) {c};
    \node[classic] (d) at (6,1) {d};
    \node[classic] (e) at (6,-2) {e};
    \node[classic] (f) at (8,2) {f};
    \node[classic] (g) at (8,0) {g};
    \node[classic] (h) at (8,-2) {h};
    \node[classic] (T) at (10,0) {T};
    
    \node[fill=white] (Inf) at (5.5,3.25) {15.462};
    
    \draw[edge_style,->,line width=0.36pt]  (S) edge node[sloped]{1.630} (a);
    \draw[edge_style,->,line width=1pt]  (S) edge node{4.559} (b);
    \draw[edge_style,->,line width=0.08pt] (S) edge node{0.349} (c);
    \draw[edge_style,->,line width=0.09pt] (a) edge node{0.431} (b);
    \draw[edge_style,->,line width=0.26pt]  (a) edge node[sloped]{1.199} (d);
    \draw[edge_style,->,line width=0.71pt]  (b) edge node[sloped]{3.224} (d);
    \draw[edge_style,->,line width=0.39pt] (b) edge node[sloped]{1.766} (e);
    \draw[edge_style,->,line width=0.08pt] (c) edge node[sloped]{0.349} (e);
    \draw[edge_style,->,line width=0.28pt]  (d) edge node[sloped]{1.268} (f);
    \draw[edge_style,->,line width=0.41pt]  (d) edge node[sloped]{1.887} (g);
    \draw[edge_style,->,line width=0.28pt] (d) edge node[sloped]{1.268} (h);
    \draw[edge_style,->,line width=0.46pt]  (e) edge node{2.115} (h);
    \draw[edge_style,->,line width=0.06pt]  (f) edge node{0.268} (g);
    \draw[edge_style,->,line width=0.22pt] (f) edge node[sloped]{1} (T);
    \draw[edge_style,->,line width=0.47pt]  (g) edge node{2.155} (T);
    \draw[edge_style,->,line width=0.74pt]  (h) edge node[sloped]{3.383} (T);
   \draw[edge_style,dotted,->] (S) |- (1,3) |- (10,3) -- (T);
\end{tikzpicture}
    
    }}
        \end{center}
    \hfill
    \\
    \subfigure[$\theta=1$]{
\resizebox{.49\linewidth}{!}{
    
        \scriptsize
\centering
\begin{tikzpicture}[shorten >=1pt, auto, node distance=1cm, thick,
   classic/.style={circle,draw=black,font=\scriptsize\bfseries},
   edge_style/.style={draw=black},scale=0.8,transform shape]
   
       \node[classic] (S) at (1,0) {S};
    \node[classic] (a) at (3,2) {a};
    \node[classic] (b) at (3,0) {b};
    \node[classic] (c) at (3,-2) {c};
    \node[classic] (d) at (6,1) {d};
    \node[classic] (e) at (6,-2) {e};
    \node[classic] (f) at (8,2) {f};
    \node[classic] (g) at (8,0) {g};
    \node[classic] (h) at (8,-2) {h};
    \node[classic] (T) at (10,0) {T};
    
    \node[fill=white] (Inf) at (5.5,3.25) {10.069};
    
    \draw[edge_style,->,line width=0.55pt]  (S) edge node[sloped]{3.953} (a);
    \draw[edge_style,->,line width=0.97pt]  (S) edge node{6.978} (b);
    \draw[edge_style,->,line width=0.14pt] (S) edge node{1} (c);
    \draw[edge_style,->,line width=0.27pt] (a) edge node{1.956} (b);
    \draw[edge_style,->,line width=0.28pt]  (a) edge node[sloped]{1.997} (d);
    \draw[edge_style,->,line width=0.83pt]  (b) edge node[sloped]{5.970} (d);
    \draw[edge_style,->,line width=0.41pt] (b) edge node[sloped]{2.963} (e);
    \draw[edge_style,->,line width=0.14pt] (c) edge node[sloped]{1} (e);
    \draw[edge_style,->,line width=0.18pt]  (d) edge node[sloped]{1.298} (f);
    \draw[edge_style,->,line width=0.48pt]  (d) edge node[sloped]{3.434} (g);
    \draw[edge_style,->,line width=0.45pt] (d) edge node[sloped]{3.236} (h);
    \draw[edge_style,->,line width=0.55pt]  (e) edge node{3.963} (h);
    \draw[edge_style,->,line width=0.04pt]  (f) edge node{0.298} (g);
    \draw[edge_style,->,line width=0.14pt] (f) edge node[sloped]{1} (T);
    \draw[edge_style,->,line width=0.52pt]  (g) edge node{3.732} (T);
    \draw[edge_style,->,line width=1pt]  (h) edge node[sloped]{7.199} (T);
   \draw[edge_style,dotted,->] (S) |- (1,3) |- (10,3) -- (T);
\end{tikzpicture}
    
    }}
    \hfill
    \subfigure[$\theta=10$]{
\resizebox{.49\linewidth}{!}{
    
        \scriptsize
\centering
\begin{tikzpicture}[shorten >=1pt, auto, node distance=1cm, thick,
   classic/.style={circle,draw=black,font=\scriptsize\bfseries},
   edge_style/.style={draw=black},scale=0.8,transform shape]
   
       \node[classic] (S) at (1,0) {S};
    \node[classic] (a) at (3,2) {a};
    \node[classic] (b) at (3,0) {b};
    \node[classic] (c) at (3,-2) {c};
    \node[classic] (d) at (6,1) {d};
    \node[classic] (e) at (6,-2) {e};
    \node[classic] (f) at (8,2) {f};
    \node[classic] (g) at (8,0) {g};
    \node[classic] (h) at (8,-2) {h};
    \node[classic] (T) at (10,0) {T};
    
    \node[fill=white] (Inf) at (5.5,3.25) {10};
    
    \draw[edge_style,->,line width=0.53pt]  (S) edge node[sloped]{4} (a);
    \draw[edge_style,->,line width=0.93pt]  (S) edge node{7} (b);
    \draw[edge_style,->,line width=0.13pt] (S) edge node{1} (c);
    \draw[edge_style,->,line width=0.27pt] (a) edge node{2} (b);
    \draw[edge_style,->,line width=0.27pt]  (a) edge node[sloped]{2} (d);
    \draw[edge_style,->,line width=0.8pt]  (b) edge node[sloped]{6} (d);
    \draw[edge_style,->,line width=0.4pt] (b) edge node[sloped]{3} (e);
    \draw[edge_style,->,line width=0.13pt] (c) edge node[sloped]{1} (e);
    \draw[edge_style,->,line width=0.13pt]  (d) edge node[sloped]{1} (f);
    \draw[edge_style,->,line width=0.47pt]  (d) edge node[sloped]{3.500} (g);
    \draw[edge_style,->,line width=0.47pt] (d) edge node[sloped]{3.500} (h);
    \draw[edge_style,->,line width=0.53pt]  (e) edge node{4} (h);
    \draw[edge_style,->,line width=0.13pt] (f) edge node[sloped]{1} (T);
    \draw[edge_style,->,line width=0.47pt]  (g) edge node{3.500} (T);
    \draw[edge_style,->,line width=1pt]  (h) edge node[sloped]{7.500} (T);
   \draw[edge_style,dotted,->] (S) |- (1,3) |- (10,3) -- (T);
\end{tikzpicture}
    
    }
    }    
    
    \caption{Representation of the net flows from $s$ to $t$ depending on the value of $\theta$, for the capacity-constrained RSP and the graph appearing in Fig.\ \ref{fig:GraphUndirected}. The thickness of the arrows is scaled with respect to the largest net flow present in the graph.
    }
    \label{fig:EvNetFlow}
\end{figure}

Fig.\  \ref{fig:EvMaxflow} shows the evolution of the flow between the nodes $\{ f,g,h \}$ and $t$ (the flow through the original graph), provided by Algorithm \ref{Alg_constrained_RSP01}, and thus satisfying the capacity constraints in terms of the value of the $\theta$ parameter. As observed in Fig.\  \ref{fig:EvMaxflow} and Fig.\ \ref{fig:EvNetFlow}, this flow through the original graph reaches almost exactly the maximum flow value of 12 for all values of $\theta$ larger than 1. However, when $\theta$ is low (close to zero), the walks became increasingly random, and no longer consider costs (see Eq.\ \ref{Eq_optimization_problem_BoP01}). For that reason, part of the total flow went through the shortcut, even if this was not optimal in terms of cost. This explains the reduction of the flow through the original graph when $\theta$ was close to zero.

\subsection{Scalability experiment}

In this subsection, we study the scalability of Algorithm \ref{Alg_randomized_shortest_path_net_dissimilarity01}, which computes the full net flow RSP dissimilarity matrix (see Eq.\ \ref{Eq_RSP_dissimilarity_definition01}), through a small experiment. 
To evaluate the time complexity, we generated 120 graphs with the benchmark algorithm of Lancichinetti, Fortunato and Radicchi (LFR) \cite{lancichinetti-2008}. More precisely, we created 10 graphs for each of the following sizes of \{50, 100, 150, 200, 250, 300, 500, 1,000, 1,500, 2,000, 2,500, 3,000\} nodes. Moreover, for each size among the 10 graphs, we changed the mixing parameter value ($\mu$) each 2 graphs between all these values \{0.1, 0.2, 0.3, 0.4, 0.5\}. 

\begin{figure}[t]
    \centering
    \subfigure{\includegraphics*[width=0.49\linewidth]{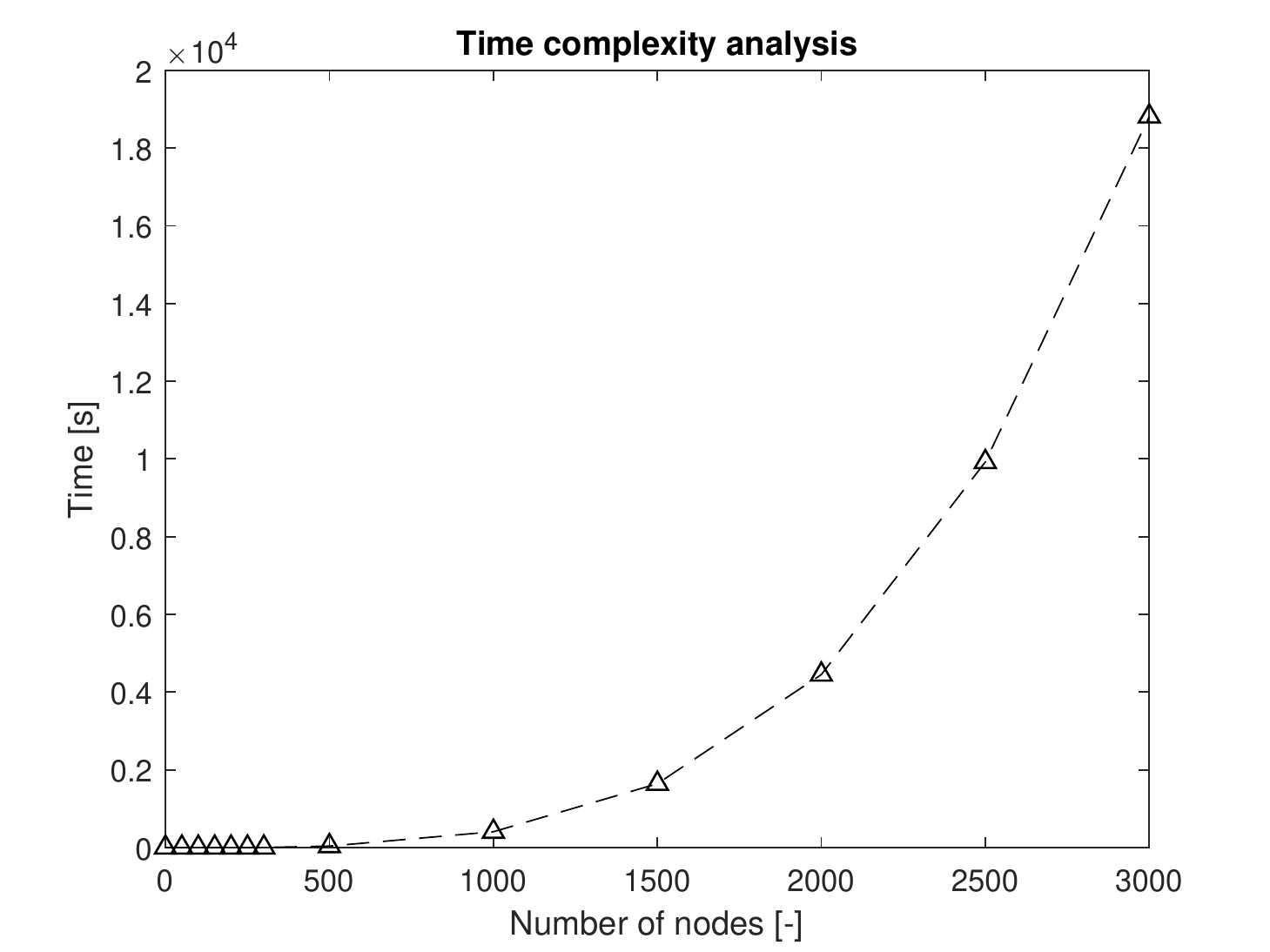}}
    \subfigure{\includegraphics*[width=0.49\linewidth]{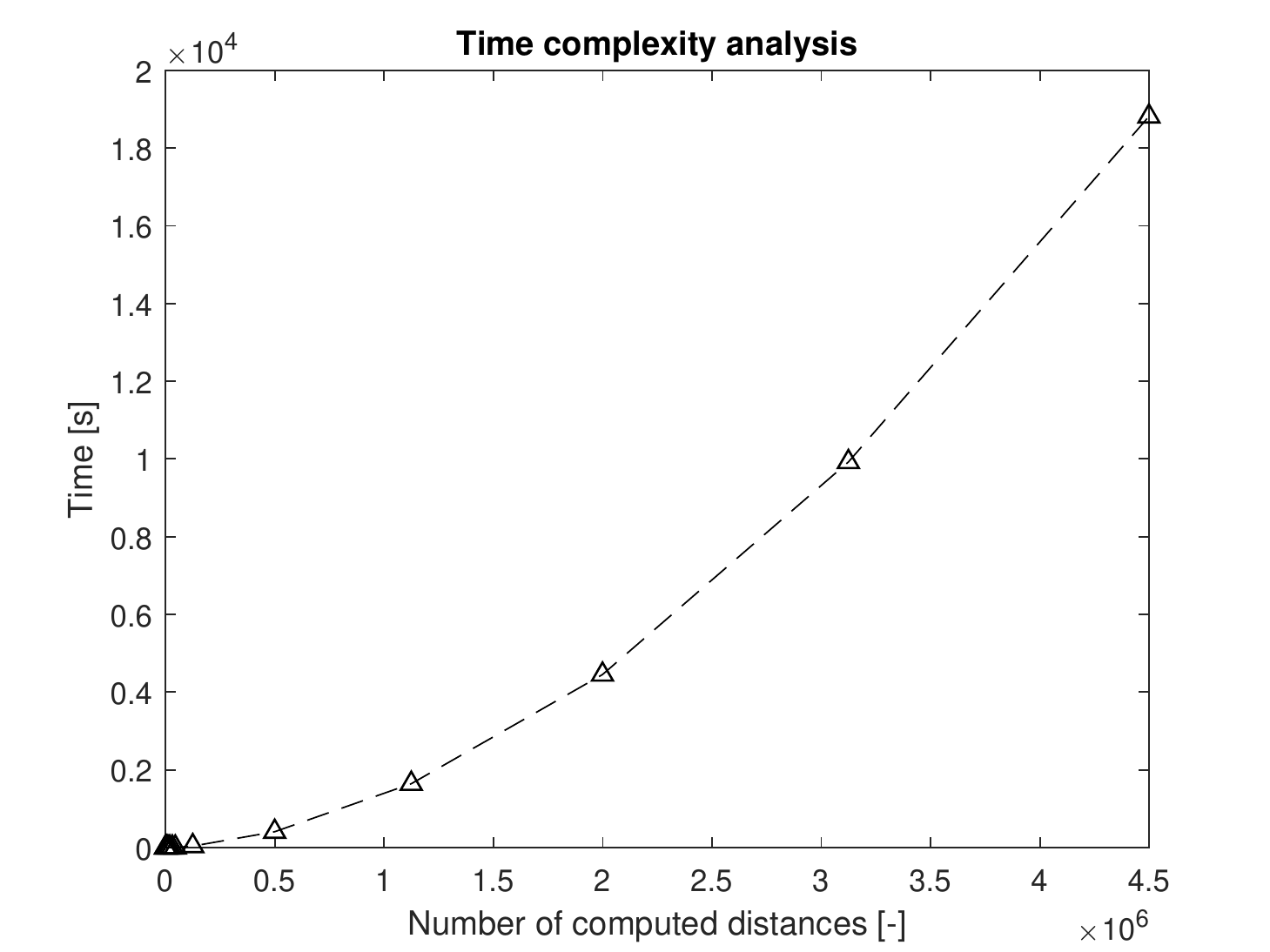}}
    \caption{Average computation time in seconds of the net flow RSP dissimilarity matrix over the 10 LFR graphs for each different network size, in terms of number of nodes (left) and number of computed dissimilarities (right).}
    \label{fig:complexity}
\end{figure}

Algorithm \ref{Alg_randomized_shortest_path_net_dissimilarity01}, computing the dissimilarity matrix between all pairs of nodes, was then run on each of these graphs. We report the average computation time in seconds over the 10 graphs for each different size (in terms of number of nodes and number of computed dissimilarities) in Fig.\ \ref{fig:complexity}. All results were obtained with Matlab (version R2017a) running on an Intel Core i7-8750H CPU@4.10GHz with 32 GB of RAM.

As previously mentioned, the time-complexity of this algorithm is $\mathcal{O}(n^3 + m n^2)$ with $n$ being the number of nodes and $m$ being the number of edges. Although applicable to medium-size graphs, it can clearly be observed that the algorithm does not scale well on large graphs in its present form, at least if the full dissimilarity matrix is needed.

\subsection{Nodes clustering experiment}
 
In this subsection, we present an application of the \textbf{Net Flow Randomized Shortest Paths} (nRSP) in a graph nodes clustering context. Note that a methodology close to \cite{Sommer-2016} is used (for further details, see the cited paper).

\subsubsection{Experimental setup}
\paragraph{Baselines}
As part of the node clustering experiment, four dissimilarity matrices between nodes as well as five kernels on a graph were used as baselines to assess our nRSP method.

\vspace*{1\baselineskip}
\noindent \textbf{Baseline dissimilarities}
\begin{itemize}
    \item The \textbf{Free Energy} (FE) \textbf{distance} and the \textbf{RSP} \textbf{dissimilarity}, depending on an inverse temperature parameter $\theta = 1/T$. As presented earlier, these methods have been shown to perform well in a node clustering context \cite{Sommer-2016} as well as in semi-supervised node classification tasks.
    \item The \textbf{Logarithmic Forest} (LF) \textbf{distance}. Introduced in \cite{Chebotarev-2011}, the LF distance relies on the matrix forest theorem \cite{Chebotarev-1997} and defines a family of distances interpolating (up to a scaling factor) between the shortest path distance and the resistance distance \cite{Klein-1993}, depending on a parameter $\alpha$.
    \item The \textbf{Shortest Path} (SP) \textbf{distance}. This well-known, standard, distance corresponds to the cost along the least-cost path between two nodes $i$ and $j$, derived from the cost matrix $\mathbf C$.
\end{itemize}
These dissimilarity matrices were transformed into inner products (kernel matrices) by classical multidimensional scaling (see below).

\vspace*{1\baselineskip}
\noindent \textbf{Baseline kernels on a graph}
\begin{itemize}
    \item The \textbf{Neumann kernel} \cite{Fouss-2016,Scholkopf-2002} (Katz) is defined as $\mathbf K=(\mathbf I-\alpha\mathbf A)^{-1}-\mathbf I$. The $\alpha$ parameter has to be positive and smaller than the inverse of the spectral radium of $\mathbf A$, $\rho(\mathbf A)=\max_i(|\lambda_i|)$.
    \item The \textbf{Logarithmic Communicability} (lCom) \textbf{kernel}, proposed in \cite{ivashkin-2016}. The lCom kernel corresponds to the logarithmic version of the communicability measure \cite{Estrada-2008}, computed as $\mathbf K=\ln(\expm{(t\mathbf A)}),\,t>0$, where $\expm$ is the matrix exponential and $\ln$ the natural elementwise logarithm (see \cite{ivashkin-2016} for details).
    \item The \textbf{Sigmoid Commute Time} (SCT) \textbf{kernel}. Proposed in \cite{Yen-2009}, the SCT kernel is obtained by applying a sigmoid transform \cite{Scholkopf-2002} on the commute-time kernel \cite{FoussKDE-2005}.  An alternative version, the \textbf{Sigmoid Corrected Commute Time} (SCCT) \textbf{kernel}, based on the correction of the commute time suggested in \cite{vonLuxburg-2010}, was also used as part of the experiments. For both methods, the parameter $\alpha$ controls the sharpness of the sigmoid. 
\end{itemize}

In addition, the \textbf{Modularity matrix} $\mathbf Q$ (Q) was used as the final baseline, and was computed by $\mathbf Q = \mathbf A-\frac{\mathbf{dd}^{\text T}}{\mathrm{vol}}$ where $\mathbf{d}$ contains the node degrees and $\mathrm{vol}$ is a constant denoting the volume of the graph (see, e.g., \cite{Newman-2018} and references therein). Recall that modularity is an unsupervised measure of the quality of a partition of the nodes (a set of communities). Here, the kernel $k$-means is executed directly on matrix $\mathbf Q$.

\paragraph{Datasets}
A collection of 17 datasets was investigated for the experimental comparisons of the dissimilarity measures. For each dataset, costs were computed as the reciprocal of affinities, $c_{ij} = 1/a_{ij}$, as in electric networks. The collection included Zachary's karate club \cite{Zachary1977}, the Dolphin datasets \cite{Lusseau-2003-emergent}, the Football dataset \cite{Newman2002}, the Political books,\footnote{Collected by V. Krebs and labelled by M. Newman, this database is not, to the best of our knowledge, published but it is available for download at \url{http://www-personal.umich.edu/~mejn/netdata/}.} three LFR benchmarks \cite{lancichinetti-2008} and nine Newsgroup datasets processed from the original Newsgroup data\footnote{Available from the UCI Machine Learning Repository.} (see \cite{Yen-2009} for details).
The list of datasets along with their main characteristics are presented in Table \ref{tab:datasets}.

\begin{table}[t]
\centering
\footnotesize
\begin{tabular}{l|l|l|l}
\hline
\textbf{Dataset Name}         & \mypound \textbf{Clusters} & \mypound \textbf{Nodes} & \mypound \textbf{Edges} \\ \hline
Dolphin\_2      & 2        & 62    & 159     \\ \hline
Dolphin\_4      & 4        & 62    & 159     \\ \hline
Football        & 12       & 115   & 613    \\ \hline
LFR1            & 3        & 600   & 6142    \\ \hline
LFR2            & 6        & 600   & 4807    \\ \hline
LFR3            & 6        & 600   & 5233   \\ \hline
Newsgroup\_2\_1 & 2        & 400   & 33854   \\ \hline
Newsgroup\_2\_2 & 2        & 398   & 21480   \\ \hline
Newsgroup\_2\_3 & 2        & 399   & 36527   \\ \hline
Newsgroup\_3\_1 & 3        & 600   & 70591   \\ \hline
Newsgroup\_3\_2 & 3        & 598   & 68201   \\ \hline
Newsgroup\_3\_3 & 3        & 595   & 64169   \\ \hline
Newsgroup\_5\_1 & 5        & 998   & 176962   \\ \hline
Newsgroup\_5\_2 & 5        & 999   & 164452  \\ \hline
Newsgroup\_5\_3 & 5        & 997   & 155618  \\ \hline
Political books & 3        & 105   & 441     \\ \hline
Zachary         & 2        & 34    & 78      \\ \hline
\end{tabular}
\caption{Datasets used in our experiments.}
\label{tab:datasets}
\end{table}
\paragraph{Evaluation metrics}
Each partition provided by an investigated clustering technique was assessed by comparing it with the ``observed partition" of the dataset. Two criteria were used to evaluate the similarity between both partitions.
\begin{itemize}
    \item The \textbf{Normalized Mutual Information (NMI)} \cite{Strehl-2002} between two partitions $\mathcal{U}$ and $\mathcal{V}$ was computed by dividing the mutual information \cite{Cover-2006} between the two partitions by the average of the respective entropy of $\mathcal{U}$ and $\mathcal{V}$.
    \item The \textbf{Adjusted Rand Index (ARI)} \cite{Hubert-1985} is an extension of the Rand Index (RI), which measures the degree of matching between two partitions. The RI has the drawback of not showing a constant expected value when working with random partitions. In contrast, the ARI has an expected value of 0 and a maximum value of 1.
\end{itemize}

\subsubsection{Experimental methodology}
The experiments relied on the kernel $k$-means introduced in \cite{Yen-2009}. For the dissimilarities, the experimental methodology was similar to the one used in \cite{Sommer-2016}. For each given dataset, the dissimilarity matrix $\mathbf D$ obtained by the different methods providing dissimilarities\footnote{For methods that directly provide a kernel, the obtained kernel matrix was directly used in the kernel k-means.} was transformed into a kernel $\mathbf K$ (a inner product matrix) using classical multidimensional scaling \cite{Fouss-2016}. If the resulting kernel was not positive semi-definite, we simply set the negative eigenvalues to zero when computing the kernel.

The kernel $k$-means was run 30 times (trials) on $\mathbf K$ with different initializations. The NMI and ARI were then computed for the partition to maximize the modularity among these 30 trials.
This operation was repeated 30 times (leading to a total of 900 runs of the $k$-means) to obtain the average modularity, and the NMI and ARI scores over these 30 repetitions for a given method (dissimilarity/similarity matrix), with a given value of its parameter (for instance, $\theta$ in the case of methods based on RSP), on a specific dataset. Finally, the reported NMI and ARI scores for each method and dataset were the average (over the 30 repetitions) for the parameter value showing the largest modularity.
Thus, modularity (instead of a separate dataset in \cite{Sommer-2016}) was used as a metric to tune the parameters of the algorithms. These parameters were the $\theta$ for the nRSP, the FE distance and the RSP dissimilarity, the $\alpha$ for the LF distance and Katz, the $t$ for the lCom kernel, and the $\alpha$ for the sigmoid transform of the SCT and SCCT kernels. The values tested for the parameter are listed in Table \ref{tab:ParamTuning}.

\begin{table}[t]
\footnotesize
\centering
\begin{tabular}{|l|l|}
\hline
\textbf{Algorithm}                                               & \textbf{Parameter values} \\ \hline
\begin{tabular}[c]{@{}l@{}}FE\\ RSP\\ nRSP\end{tabular} & $\theta= (0.001,\,0.005,\,0.01,\,0.05,\,0.1,\,0.5,\,1,\,3,\,5,\,10,\,15,\,20)$\\ \hline
LF                                                      & $\alpha= (0.001,\,0.005,\,0.01,\,0.05,\,0.1,\,0.5,\,1,\,3,\,5,\,10,\,15,\,20)$\\ \hline
Katz                                                    & $\alpha=(0.05,\,0.10,\dots,\,0.95) \times (\rho(\mathbf A))^{-1}$\\ \hline
lCom                                                    & $t=(0.01,\,0.02,\,0.05,\,0.1,\,0.2,\,0.5,\,1,\,2,\,5,\,10)$\\ \hline
\begin{tabular}[c]{@{}l@{}}SCT\\ SCCT\end{tabular}      & $\alpha=(5,\,10,\,15,\,\dots,\,50)$\\ \hline
\end{tabular}
\caption{Parameter range for the investigated methods.}
\label{tab:ParamTuning}
\end{table}

\subsubsection{Experimental results}

\begin{figure}[t]
    \centering
    \subfigure{\includegraphics*[width=0.49\linewidth,trim= 0 20 0 0]{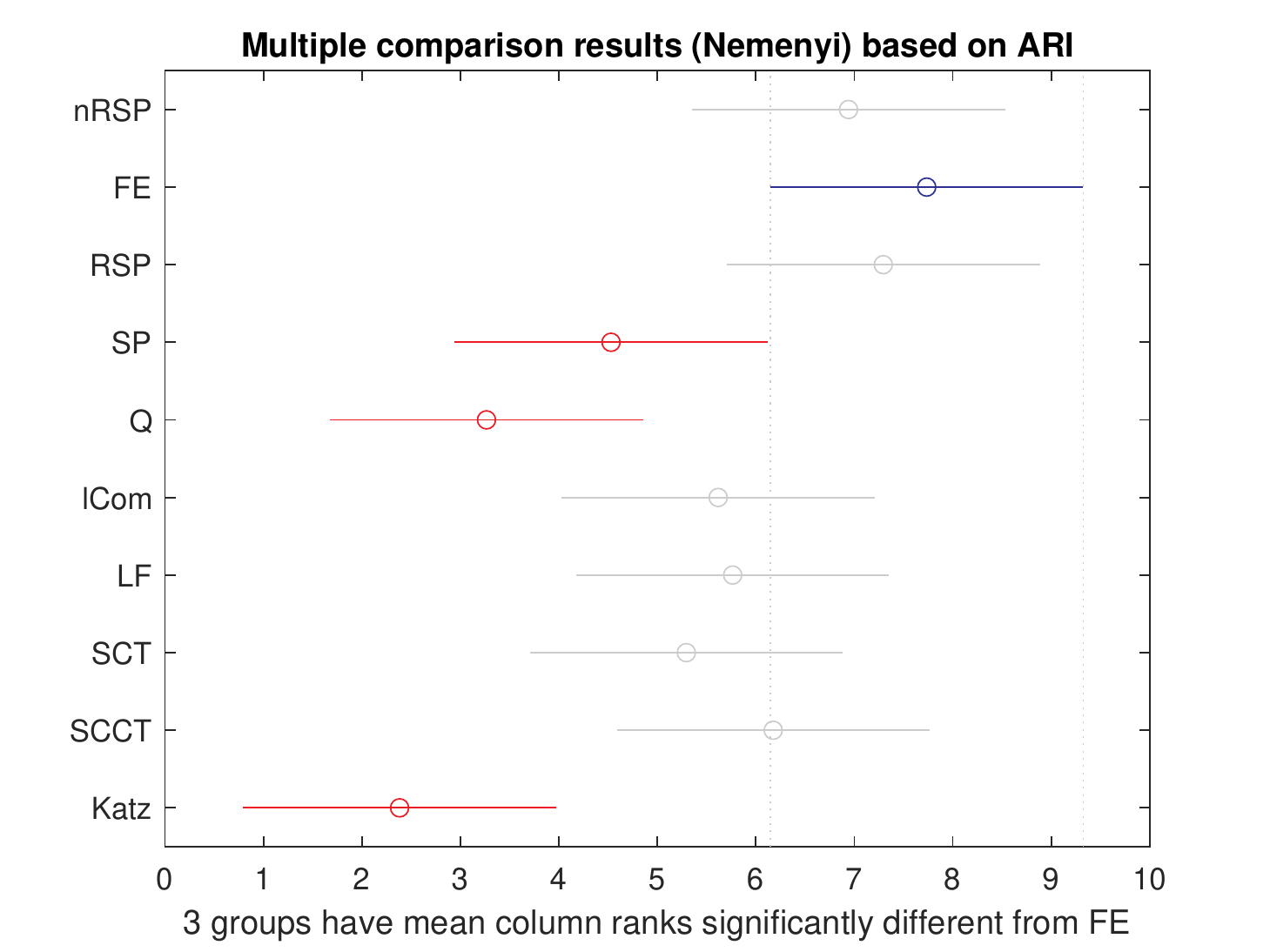}}
    \subfigure{\includegraphics*[width=0.49\linewidth,trim= 0 20 0 0]{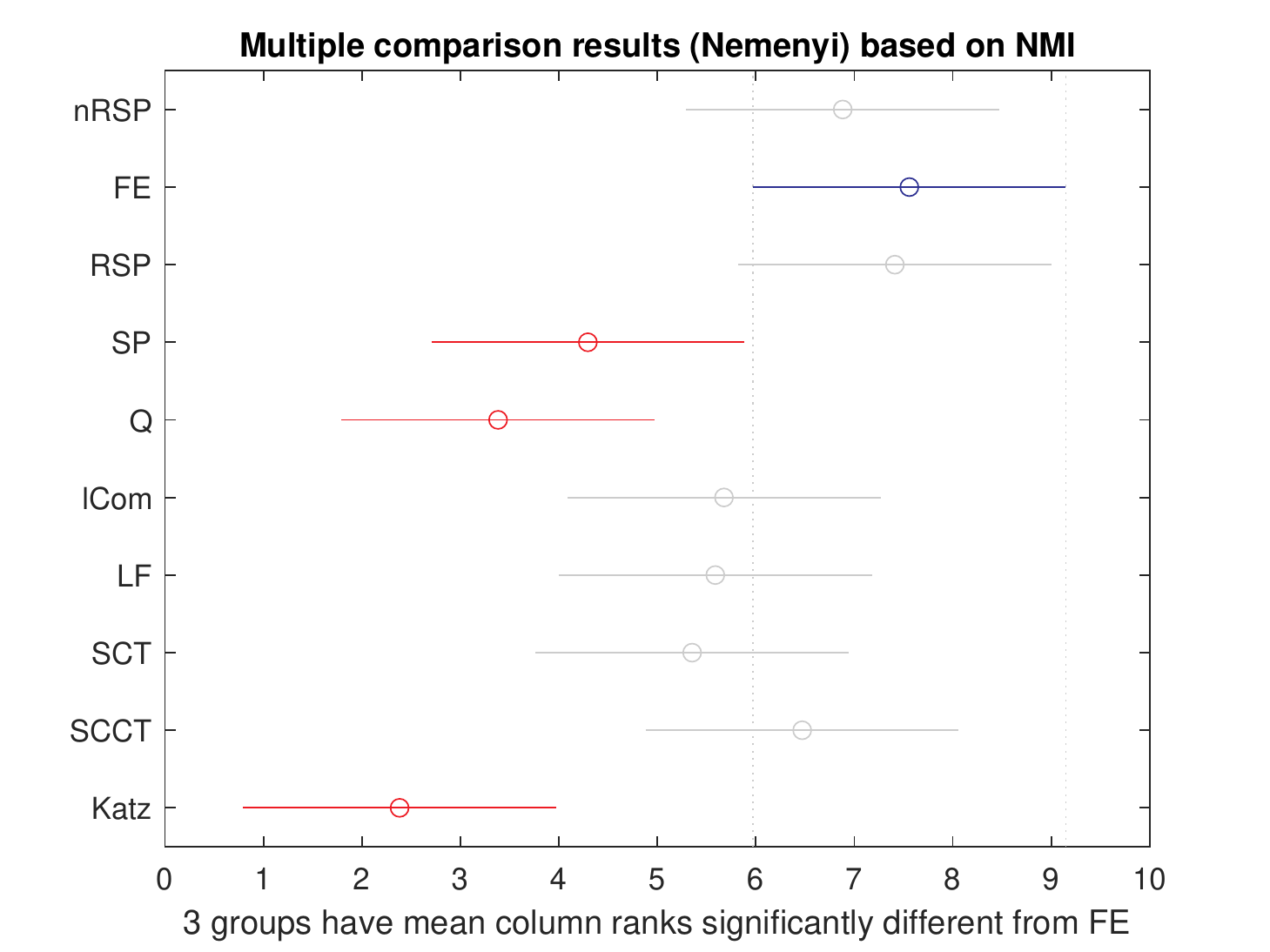}}
    \caption{Mean ranks and 95\% Nemenyi confidence intervals of the tested methods across the 17 datasets for the ARI (left) and the NMI (right) measures. The larger the rank, the better.}
    \label{fig:nemenyiclustering}
\end{figure}

The different methods were globally assessed using the same method as in \cite{Sommer-2016} based on a non-parametric Friedman-Nemenyi test \cite{Demvsar-2006}. Additionally, a non-parametric Wilcoxon signed-rank test was performed pairwise to measure the significance of the difference in the algorithms' performance. 

\begin{table}[t]
\centering
\scriptsize
\begin{tabular}{|c|c|c|c|c|c|c|c|c|c|c|}
\hline
\textbf{Method} & nRSP  & FE    & RSP   & SP    & Q     & lCom  & LF    & SCT   & SCCT  & Katz  \\ \hline
nRSP   & \cellcolor[HTML]{000000}1     &0.389& 0.497 &\cellcolor[HTML]{C0C0C0}\textbf{0.017}& \cellcolor[HTML]{C0C0C0}\textbf{0.004} & 0.241 & 0.188 & 0.151 & 0.561 & \cellcolor[HTML]{C0C0C0}\textbf{0.001} \\ \hline
FE     & 0.277 & \cellcolor[HTML]{000000}1     & 0.626 & \cellcolor[HTML]{C0C0C0}\textbf{0.015} & \cellcolor[HTML]{C0C0C0}\textbf{0.002} & 0.068 & \cellcolor[HTML]{C0C0C0}\textbf{0.025} & \cellcolor[HTML]{C0C0C0}\textbf{0.015} & 0.127 & \cellcolor[HTML]{C0C0C0}\textbf{0.001} \\ \hline
RSP    & 0.588 & 0.153 & \cellcolor[HTML]{000000}1     & \cellcolor[HTML]{C0C0C0}\textbf{0.011} &\cellcolor[HTML]{C0C0C0}\textbf{0.002} & \cellcolor[HTML]{C0C0C0}\textbf{0.025} & \cellcolor[HTML]{C0C0C0}\textbf{0.020} & \cellcolor[HTML]{C0C0C0}\textbf{0.026 }& 0.194 & \cellcolor[HTML]{C0C0C0}\textbf{0.001} \\ \hline
SP     & \cellcolor[HTML]{C0C0C0}\textbf{0.019} & \cellcolor[HTML]{C0C0C0}\textbf{0.015} & \cellcolor[HTML]{C0C0C0}\textbf{0.031} & \cellcolor[HTML]{000000}1     & 0.246 & \cellcolor[HTML]{C0C0C0}\textbf{0.028} & 0.093 & 0.062 & \cellcolor[HTML]{C0C0C0}\textbf{0.011} & \cellcolor[HTML]{C0C0C0}\textbf{0.004} \\ \hline
Q      & \cellcolor[HTML]{C0C0C0}\textbf{0.002}& \cellcolor[HTML]{C0C0C0}\textbf{0.001} & \cellcolor[HTML]{C0C0C0}\textbf{0.001} & 0.055 & \cellcolor[HTML]{000000}1     & \cellcolor[HTML]{C0C0C0}\textbf{0.006} & \cellcolor[HTML]{C0C0C0}\textbf{0.005} & \cellcolor[HTML]{C0C0C0}\textbf{0.004 }& \cellcolor[HTML]{C0C0C0}\textbf{0.002} & 0.076 \\ \hline
lCom   & 0.241 & \cellcolor[HTML]{C0C0C0}\textbf{0.042} & 0.078 & 0.068 & \cellcolor[HTML]{C0C0C0}\textbf{0.004} & \cellcolor[HTML]{000000}1     & 0.855 & 1.000     & 0.241 & \cellcolor[HTML]{C0C0C0}\textbf{0.001} \\ \hline
LF     & 0.151 & \cellcolor[HTML]{C0C0C0}\textbf{0.011} & 0.078 & 0.062 & \cellcolor[HTML]{C0C0C0}\textbf{0.003} & 0.952 & \cellcolor[HTML]{000000}1     & 0.934 & 0.217 & \cellcolor[HTML]{C0C0C0}\textbf{0.001} \\ \hline
SCT    & 0.055 & \cellcolor[HTML]{C0C0C0}\textbf{0.012} & \cellcolor[HTML]{C0C0C0}\textbf{0.030}  & 0.163 & \cellcolor[HTML]{C0C0C0}\textbf{0.004} & 0.359 & 0.359 & \cellcolor[HTML]{000000}1     & \cellcolor[HTML]{C0C0C0}\textbf{0.009} &\cellcolor[HTML]{C0C0C0}\textbf{0.001} \\ \hline
SCCT   & 0.252 & \cellcolor[HTML]{C0C0C0}\textbf{0.048} & 0.153 & \cellcolor[HTML]{C0C0C0}\textbf{0.044} &\cellcolor[HTML]{C0C0C0}\textbf{0.002} & 0.808 & 0.426 & \cellcolor[HTML]{C0C0C0}\textbf{0.004} &\cellcolor[HTML]{C0C0C0} \cellcolor[HTML]{000000}1     & \cellcolor[HTML]{C0C0C0}\textbf{0.001}\\ \hline
Katz   & \cellcolor[HTML]{C0C0C0}\textbf{0.001} & \cellcolor[HTML]{C0C0C0}\textbf{0.001} & \cellcolor[HTML]{C0C0C0}\textbf{0.001} & \cellcolor[HTML]{C0C0C0}\textbf{0.002} & 0.062 & \cellcolor[HTML]{C0C0C0}\textbf{0.001} & \cellcolor[HTML]{C0C0C0}\textbf{0.001} & \cellcolor[HTML]{C0C0C0}\textbf{0.001} & \cellcolor[HTML]{C0C0C0}\textbf{0.001} & \cellcolor[HTML]{000000}1     \\ \hline
\end{tabular}
\caption{The $p$-values provided by a pairwise Wilcoxon signed-rank test, for the NMI in the upper right triangle and the ARI in the lower left.}
\label{tab:wcx}
\end{table}

The results of the Friedman-Nemenyi test are summarized in Fig.\ \ref{fig:nemenyiclustering}. Additionally, Table \ref{tab:wcx} contains the pairwise $p$-values for the Wilcoxon signed-rank test performed on the results obtained from the 17 datasets. More specifically, the upper-right side of the diagonal of the matrix contains the $p$-values when considering NMI and the lower-left side contains the $p$-values when using the ARI.

We can observe on the Nemenyi plot that the three leading methods appear to be the FE, the RSP and the nRSP. More specifically, the FE was the best method on the investigated datasets and in this setup. Based on the Wilcoxon test using ARI, the FE performed significantly better than all the other methods, at a $\alpha = 0.05$ level, except for the RSP and nRSP. However, note that, for the NMI score, the difference between the FE and the lCom and between the FE and the SCCT were not significant.

The introduced method (nRSP) obtained results comparable to the RSP and the FE for both the ARI and the NMI measures. None of these differences were significant after performing a Wilcoxon signed rank test ($\alpha = 0.05$). Although not the best overall, the introduced nRSP method proved to be competitive with respect to the FE and the RSP, which, in turn, performed best in a similar but more extensive node clustering comparison \cite{Sommer-2016}. It can also be observed that the nRSP and the standard RSP performed very similarly, which was somewhat expected because both dissimilarities are based on the same framework.

\section{Conclusion}
\label{Sec_conclusion01}

In this work, two extensions of the RSP formalism were developed. The first extension introduces an algorithm for computing the expected net costs between all pairs of nodes by considering the \emph{net flows} between nodes instead of the raw flows. This quantity is called the net flow RSP dissimilarity; it quantifies the level of accessibility (proximity and ease of access) between nodes \cite{Chebotarev-1997} and serves as a dissimilarity measure. 
The second extension deals with \emph{capacity constraints} on edges for both raw and net flows within the RSP formalism. An algorithm solving the constrained problem has been developed.

These contributions extend the scope of the RSP formalism, which essentially defines a model of movement, or spread, through the network. Indeed, as already discussed in the introduction, many of the traditional models are based on two common paradigms about the transfer of information, or more generally the movement, occurring in the network: an optimal behavior based on least-cost paths and a random behavior based on a random walk on the graph.
In contrast to these standard models, the RSP, and other families of distances (see the related work in the introductory Section \ref{Sec_introduction01}), interpolate between a pure random walk on the graph and an optimal behavior based on shortest paths. They depend on a parameter that allows the amount of randomness of the trajectories to be monitored. This acknowledges the fact that in many practical cases the (random) walker on the graph is neither completely rational nor completely stochastic.

Another peculiarity of the RSP model is that it adopts a statistical physics framework that considers the system of all paths (or walks) connecting pairs of nodes in the network. The path-based formalism assigns a Gibbs-Boltzmann probability distribution on the set of paths by minimizing expected cost with relative entropy regularization weighted by temperature -- the free energy objective function. 

Different quantities capturing the degree of relative accessibility between the nodes can then be derived within this formalism, showing different properties depending on the temperature controlling randomness. Another interesting feature is that most quantities of interest can be computed in closed form by using standard matrix operations.

Experimental comparisons on clustering tasks demonstrated that the net flow RSP dissimilarity is competitive in comparison with other state-of-the-art baseline methods. This indicates that the model is able to capture the cluster structure of networks in an accurate way on the investigated datasets. Indeed, based on the same framework, the net flow RSP obtained results comparable to the simple RSP and the free energy dissimilarities.

In conclusion, the contributions of this paper should enlarge the range of possible applications of the RSP formalism. Indeed, many problems related to the spread of information in a network involve capacity constraints on edges. Moreover, in many real cases, it can be argued that a model based on unidirectional net flows is more realistic than raw flows going back and forth.

Concerning further work, we are also interested in applying the proposed models to operations research problems. Indeed, it would certainly be interesting to compare the RSP solution (with entropy regularization) to more standard algorithms for solving minimum cost flow problems and minimum cost flow with capacity constraints problems \cite{Ahuja-1993,Korte-2018}. 

In addition, more sophisticated optimization techniques, going beyond the simple gradient technique used for solving the capacity-constrained RSP problem in Section \ref{Sec_edge_flow_capacity_constraints01}, should be investigated. Still another interesting idea would be to try to reformulate the optimization problem in terms of expected net costs instead of expected costs in Eq.\ \ref{Eq_Lagrange_node_net_flow_constraints_inequality01}.

We also plan to explore a route that could improve the scalability of the proposed algorithms on sparse graphs. Following \cite{Dial71}, the idea would be to extract a DAG from the original $s$-$t$ graph by, for instance, performing a breath-first traversal or computing the electric current flow between the source node $s$ and the target node $t$. Recall that we saw in Subsection \ref{Subsec_electrical_current_DAG01} that the electric current defines a DAG. It should, therefore, be possible to compute efficiently the optimal policy (and, consequently, the directed flows) on this DAG by using the Bellman-Ford-like expression that computes the free energy directed distance (see \cite{Fouss-2016}). It would also be interesting to explore the introduction of capacity constraints on a DAG, as already discussed in Subsection \ref{Subsec_resulting_algorithm_net_flows01}.

\begin{center}
\rule{2.5in}{0.01in}
\end{center}

\section*{Acknowledgements}

\noindent This work was partially supported by the Immediate and the Brufence projects funded by InnovIris (Brussels region), as well as former projects funded by the Walloon region, Belgium. We thank these institutions for giving us the opportunity to conduct both fundamental and applied research. Moreover, we thank Professor Bernard Fortz (Universit\'e Libre de Bruxelles) for his remarks on the RSP formalism. Finally, we are also grateful to the anonymous reviewers whose suggestions have helped us significantly to improve the manuscript. Marco Saerens is also 'Collaborateur scientifique' at IRIDIA lab, Université Libre de Bruxelles (ULB).

\bibliographystyle{myabbrv}
\bibliography{Biblio.bib}

\begin{center}
\rule{2.5in}{0.01in}
\end{center}

\end{document}